\documentclass[nohyperref]{article}

\usepackage{microtype}
\usepackage{graphicx}
\usepackage{booktabs} 

\usepackage{hyperref}

 \usepackage[accepted]{icml2022}

\usepackage{amsmath}
\usepackage{amssymb}
\usepackage{mathtools}
\usepackage{amsthm}

\usepackage[capitalize,noabbrev]{cleveref}
\usepackage{subcaption}
\usepackage{multirow}
\usepackage{booktabs}
\usepackage[normalem]{ulem}
\usepackage{bbm}
\usepackage{custom_math}
\usepackage{placeins}
\usepackage[short,nocomma]{optidef}
\usepackage{soul}
\usepackage{enumitem}
\usepackage{booktabs}
\usepackage{tabu}

\usepackage{algpseudocode}

\makeatletter
\newcommand{\ssymbol}[1]{^{\@fnsymbol{#1}}}
\makeatother

\newcommand{\LL}[1]{$\ell_{#1}$}

\newcommand{\ModelLR}{f}
\newcommand{\ModelHR}{F}
\newcommand{\Scale}{g}
\newcommand{\Defense}{h}
\newcommand{\SpaceLR}{\mathbb{L}}
\newcommand{\SpaceHR}{\mathbb{H}}
\newcommand{\Kernel}{\bm{k}}

\theoremstyle{plain}

\theoremstyle{definition}

\theoremstyle{remark}

\usepackage[textsize=normalsize,prependcaption]{todonotes}

\icmltitlerunning{The Interplay Between Vulnerabilities in Machine Learning Systems}

\begin{document}

\twocolumn[
\icmltitle{Rethinking Image-Scaling Attacks: \\
The Interplay Between Vulnerabilities in Machine Learning Systems}



\icmlsetsymbol{equal}{*}

\begin{icmlauthorlist}
\icmlauthor{Yue Gao}{wisc}
\icmlauthor{Ilia Shumailov}{vector}
\icmlauthor{Kassem Fawaz}{wisc}
\end{icmlauthorlist}

\icmlaffiliation{wisc}{University of Wisconsin--Madison, Madison, WI, USA}
\icmlaffiliation{vector}{Vector Institute, Toronto, ON, Canada}

\icmlcorrespondingauthor{Yue Gao}{gy@cs.wisc.edu}


\icmlkeywords{Machine Learning, ICML}

\vskip 0.3in
]



\printAffiliationsAndNotice{}  

\begin{abstract}

As real-world images come in varying sizes, the machine learning model is part of a larger system that includes an upstream image scaling algorithm. In this paper, we investigate the interplay between vulnerabilities of the image scaling procedure and machine learning models in the decision-based black-box setting. We propose a novel sampling strategy to make a black-box attack exploit vulnerabilities in scaling algorithms, scaling defenses, and the final machine learning model in an end-to-end manner. Based on this scaling-aware attack, we reveal that most existing scaling defenses are ineffective under threat from downstream models. Moreover, we empirically observe that standard black-box attacks can significantly improve their performance by exploiting the vulnerable scaling procedure. We further demonstrate this problem on a commercial Image Analysis API with decision-based black-box attacks.

\end{abstract}

\section{Introduction}
\label{sec:introduction}

Recent advances in machine learning (ML) techniques have demonstrated human-level performance in many vision tasks, such as image classification~\cite{imagenet,inception,inception-v4} and object detection~\cite{faster-rcnn,yolo}. As real-world images come in varying sizes, the practical ML system must include an image scaling algorithm before the downstream ML model. The scaling algorithm resizes input images to match the fixed input size of a model, which takes the input and performs vision tasks, such as classification.

The model and scaling algorithm in an ML system have become attractive targets for attackers. ML models are vulnerable to adversarial examples~\cite{szegedy2013intriguing,biggio2013evasion}: an adversary can add imperceptible perturbations to the input of a model and change its prediction~\cite{pgd,carlini}. Recently, scaling algorithms were also found to be vulnerable to image-scaling attacks~\cite{scaling2019,scaling2020}: an adversary can manipulate a large image such that it will change into a different image after scaling, thereby inducing an incorrect prediction in the model.
However, image-scaling attacks are easily blocked by subsequent defenses~\cite{scaling2020,detection2020} and are not generalizable to other fields. We note that more subtle exploitation should leverage the weakness of the scaling stage \emph{jointly} with adversarial examples against the ML model.

In this paper, we investigate the interplay between vulnerabilities of the image scaling procedure and ML models. Our investigation focuses on the more practical decision-based black-box setting, where the attacker can only query the model without a confidence score or knowledge of its internal implementation. We show that the attacker can jointly attack the scaling procedure and the ML model, posing more serious threats. From one side, black-box attacks~\cite{boundaryattack,hopskip,optattack,signoptattack,hsj-linear,hsj-nonlinear,hsj-optimal} can leverage the weakness of the scaling function to improve their performance significantly. On the other side, most image-scaling defenses~\cite{scaling2020,detection2020} are not effective in protecting the scaling function from being exploited by adversarial examples, even if they successfully prevent the image-scaling attack.

\begin{figure}[tb]
	\centering
	\includegraphics[width=\linewidth]{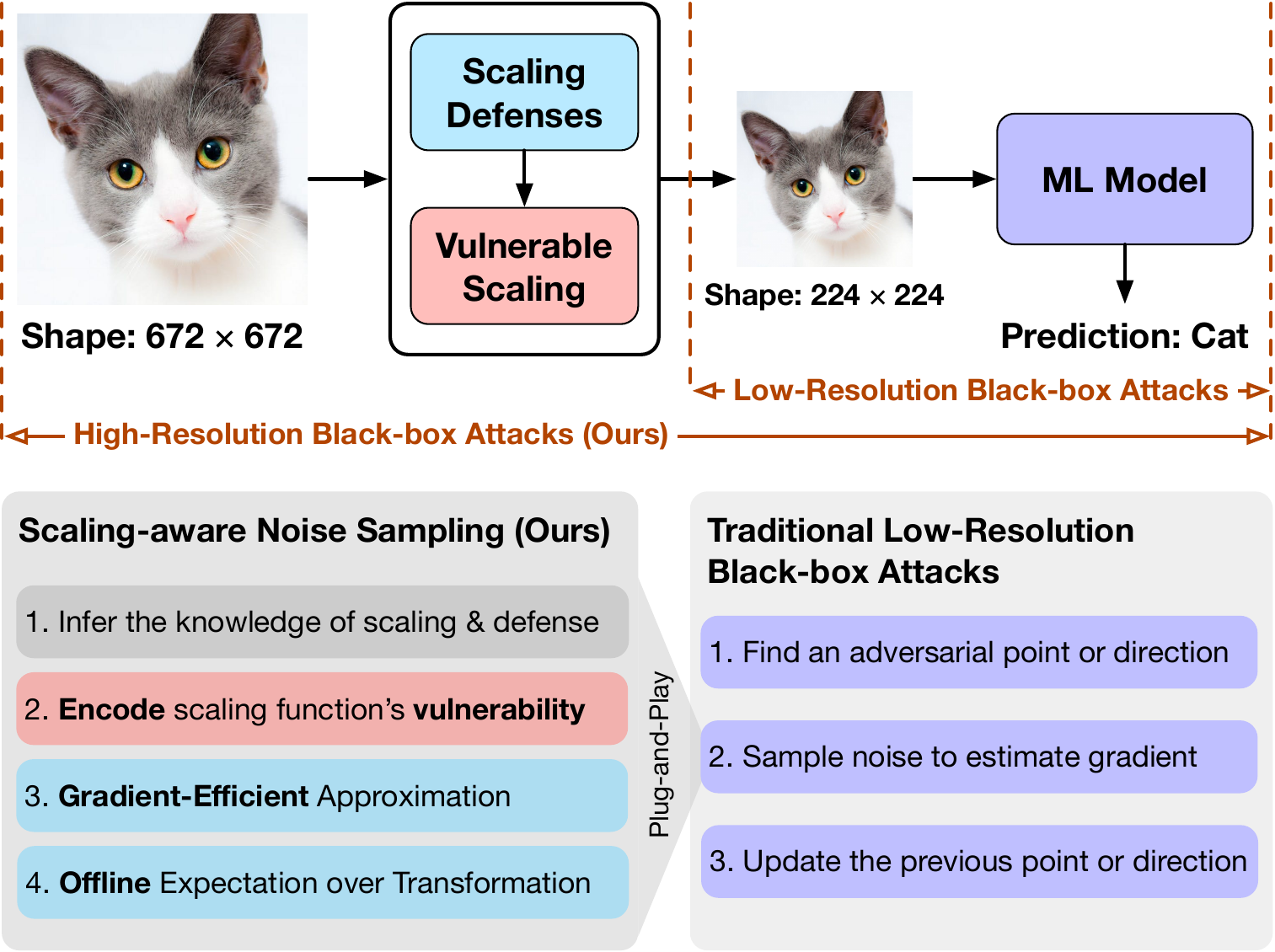}
	\caption{An overview of scaling-aware noise sampling. This plug-and-play technique guides traditional black-box attacks to exploit the weakness of the scaling function to improve their performance significantly.}
	\label{fig:demo:attack}
\end{figure}

As a first step, we generalize the attack setting and re-design black-box attacks to jointly exploit the scaling function. We characterize the common approach of existing black-box attacks and identify \emph{noise sampling} as a critical step shared by these attacks. Based on this observation, we propose to incorporate the weakness of the scaling function through a novel technique which we call Scaling-aware Noise Sampling (SNS). The overview of SNS is illustrated in \Cref{fig:demo:attack}. The high-level idea is guiding traditional \emph{low-resolution} black-box attacks to search for the adversarial examples along a direction that best exploits the scaling function. Our utilization of the noise sampling step makes SNS a plug-and-play technique applicable to different attacks. In particular, we integrate SNS with two representative decision-based black-box attacks: the boundary-based HSJ~\cite{hopskip} and the optimization-based Sign-OPT~\cite{signoptattack} attacks. We call these \textit{high-resolution attacks}, targeting the ML pipeline as a whole and exploiting scaling.

Next, we design two novel techniques to circumvent image-scaling defenses~\cite{scaling2020}. Such defenses not only protect the scaling function but also hinder the traditional black-box attacks. To incorporate these defenses in our high-resolution attacks, we identify their root mechanism and propose novel approximations. First, we design \emph{improved gradient estimation} for the defense that slows down the black-box attack's convergence with non-useful gradients. Second, we design \emph{offline} expectation over transformation~\cite{eot} to attack randomized defenses without additional queries to the black-box model. With these techniques, we circumvent 4 out of 5 state-of-the-art defenses to exploit the scaling function.

Finally, we conduct extensive experimentation to evaluate our high-resolution black-box attacks and state-of-the-art image-scaling defenses. We empirically confirm that jointly attacking the scaling function and ML model effectively improves the performance of black-box attacks. We also show that the defended scaling functions retain weaknesses that enable a stronger black-box attack. We finish with a discussion about the evaluation of trustworthy ML.

\underline{\textbf{Contributions.}}
We take the \emph{first} step towards exploring the interplay between different vulnerabilities in black-box ML systems, with additional novel insights into circumventing pre-processing defenses in the black-box setting. Our contributions can be summarized as follows.
\begin{itemize}[leftmargin=*]
	\item \textbf{Improving black-box attacks.} We propose a novel plug-and-play technique called \emph{scaling-aware noise sampling}, which significantly improves black-box attacks when a vulnerable scaling algorithm precedes the ML model.
	
	\item \textbf{Circumventing image-scaling defenses.} We show that 4 out of 5 state-of-the-art defenses, designed to protect the scaling stage, retain weaknesses that enable stronger black-box attacks.
	
	\item \textbf{New perspective of trustworthy ML.} We reveal that preventing attacks targeting one component in the ML system does not necessarily mitigate the vulnerability from a broader perspective. The interplay of different vulnerabilities leads to unexpected and stronger threats, such as amplifying existing attacks.
\end{itemize}

\section{Related Work}

Black-box attacks against ML models are drawing increasing attention due to their practical setting. In the black-box setting, the attacker can only query the target model without knowledge of its internal implementation. Despite attacks leveraging the transferability of adversarial examples~\cite{transferability}, the more common query-based attacks fall into two categories: \emph{score-based} and \emph{decision-based} attacks. Score-based attacks assume access to the confidence score of the target model's prediction, which facilitates estimating the model's gradient~\cite{soft1,soft2,soft3,soft4}. Decision-based attacks require access to the final decision without confidence scores, which is challenging but more practical~\cite{boundaryattack,hopskip,optattack,signoptattack,hsj-linear,hsj-nonlinear,hsj-optimal}. However, all these attacks have only focused on the standalone model, omitting that practical ML models must include a scaling function to cater for input images of varying sizes.

Recently, image-scaling attacks~\cite{scaling2019,scaling2020} reveal that the scaling function can be exploited to hide a small image (from a different class) into a larger image, thereby fooling the model after downscaling. However, this exploitation is easily prevented~\cite{scaling2020,detection2020} and not generalizable to other fields. This paper shows that more subtle exploitation should leverage the weakness of scaling functions to hide the adversarial perturbation rather than an unperturbed image.

Understanding the trustworthiness of ML models is a critical objective in practice. In this work, we aim to explore if the interplay between vulnerabilities in practical ML systems could cause more harm. From one side, we explore whether the vulnerability of scaling functions could amplify black-box attacks. From the other side, we explore whether black-box attacks could still exploit defended scaling functions.

\section{Background}

\subsection{Notation}
\label{sec:pre:notations}
A standard neural network $\ModelLR$ classifies low-resolution (LR) images $\bm{x}\in\SpaceLR\coloneqq[0,1]^{p\times q}$ of height $p$ and width $q$. For example, ResNet-50~\cite{resnet} accepts $224\times224$ input images\footnote{We omit the channel dimension for simplicity.}. A scaling function $\Scale$ downscales the high-resolution (HR) image $X\in\SpaceHR\coloneqq[0,1]^{m\times n}$ to the network's LR input space $\SpaceLR$. A pre-processor $\Defense$ sanitizes the input $X$ to prevent the attacker from hiding perturbation via the scaling function $\Scale$.
We focus on the black-box attack against the end-to-end classifier $\ModelHR\coloneqq \ModelLR\circ\Scale\circ\Defense$. For both $\ModelLR$ and $\ModelHR$ we denote their outputs as the classification label.
We provide more background of scaling in \Cref{app:scaling}.

\subsection{Adversarial Examples}
Given an image $\bm{x}\in\SpaceLR$ and a classifier $\ModelLR$, the traditional adversarial example $\bm{x}^\prime$ is visually similar to $\bm{x}$ but misclassified, i.e., $\ModelLR(\bm{x}^\prime)\neq \ModelLR(\bm{x})$~\cite{szegedy2013intriguing,biggio2013evasion}. Traditional attacks construct the adversarial example by searching for $\bm{\delta}$ such that $\ModelLR(\bm{x}+\bm{\delta})\neq \ModelLR(\bm{x})$, while minimizing $\norm{\bm{\delta}}$ or maximizing the loss on $\ModelLR(\bm{x}+\bm{\delta})$.

In this paper, we further consider the \emph{high-resolution} adversarial example. Given an image $X\in\SpaceHR$ and an end-to-end classifier $\ModelHR$, the \emph{high-resolution} adversarial example $X^\prime$ is visually similar to $X$ but misclassified, i.e., $\ModelHR(X^\prime)\neq\ModelHR(X)$. 
Compared to the \emph{low-resolution} $\bm{x}^\prime$, the \emph{high-resolution} $X^\prime$ additionally passes through the scaling stage $\Scale\circ\Defense$.

\subsection{Image-scaling Attacks}
\label{sec:pre:scaling-attack}
Recent image-scaling attacks~\cite{scaling2019,scaling2020} attack the scaling stage $\Scale$ to hide a smaller non-adversarial image $\bm{x}^*$ from a different class into the larger image $X$.
Conceptually, the adversarial image is visually similar to $X$ but changes into a different image $\bm{x}^*$ after scaling, thereby inducing misclassification. The details and formulation of this attack can be found in \Cref{app:scaling-attacks}.

\subsection{Image-scaling Defenses}
The image-scaling attacks are prevented by several defenses, which fall into \emph{pre-processing} and \emph{detection} defenses. We briefly introduce five state-of-the-art defenses below and provide more details in \Cref{app:scaling-defense}.

\textbf{Pre-processing Defenses.} \citet{scaling2020} propose to sanitize the input image with \emph{median} or \emph{randomized} filtering operations before scaling; they reconstruct pixels by a median or randomly picked pixel within the sliding window. These defenses instantiate the preprocessor $\Defense$ in \Cref{sec:pre:notations}. 

\textbf{Detection Defenses.} \citet{detection2020} propose three detection defenses using spatial and frequency transformations: unscaling, minimum-filtering, and centered spectrum. These transformations result in discernible differences when applied to benign and perturbed images. For example, the unscaling invokes downscale and upscale operations sequentially to reveal the hidden image.

\subsection{Threat Models}
\label{sec:pre:threat}

We focus on the decision-based black-box setting, where we do not know the implementation of $\ModelLR$ and $\ModelHR$; we only know the final predicted label without scores. This is the same threat model as considered by image-scaling attacks and defenses~\cite{scaling2020,detection2020}.
However, we can still \emph{infer} the knowledge of the scaling stage $\Scale\circ\Defense$ with the brute-forcing method from \citet{scaling2019}. This knowledge can be reused for subsequent attacks, as it is typically fixed for a deployed ML model. We explain more details of this method in \Cref{sec:background:scaling-defenses:infer}.

The objective of our attack is to leverage the weakness of the scaling function to amplify existing black-box attacks in terms of fewer queries, less perturbation, and higher optimization efficiency. Since our attack exploits the weakness of the scaling function, we also incorporate defenses that are supposed to protect it. \emph{As a result, attacking the full pipeline $\ModelHR$ (with scaling) yields significantly better performance than attacking the standalone model $\ModelLR$ (without scaling).}

\section{Attack Methodology}
\label{sec:method}

Existing black-box attacks have only focused on the model $\ModelLR$ rather than the end-to-end pipeline $\ModelHR$. Even if these attacks are deployed to attack the full pipeline $\ModelHR$, they are not designed to exploit the scaling stage $\Scale\circ\Defense$ to hide the adversarial perturbation. Therefore, the main challenge is \emph{how to make black-box attacks aware of the weakness of the scaling function in a generalizable manner}.

To address this challenge, we propose a novel technique which we call Scaling-aware Noise Sampling (SNS). As illustrated in \Cref{fig:demo:attack}, SNS is a plug-and-play technique that guides existing black-box attacks to leverage the weakness of the scaling stage explicitly. We also propose two novel designs to let SNS incorporate pre-processors that are supposed to protect the scaling function.

\subsection{Scaling-aware Noise Sampling (SNS)}
\label{sec:method:sns}

We start by characterizing the common property of existing black-box attacks. Most decision-based black-box attacks walk near the decision boundary~\cite{boundaryattack,hopskip,hsj-linear,hsj-nonlinear,hsj-optimal} or optimize for a particular objective function~\cite{optattack,signoptattack}. These attacks have varying algorithms, but all share a similar iterative approach: (1) find an adversarial point or direction through linear search; (2) sample noise to estimate the gradient of a particular objective function; and (3) use the gradient to update the previous point or direction.

\begin{figure}[tb]
    \centering
    \includegraphics[width=0.7\linewidth]{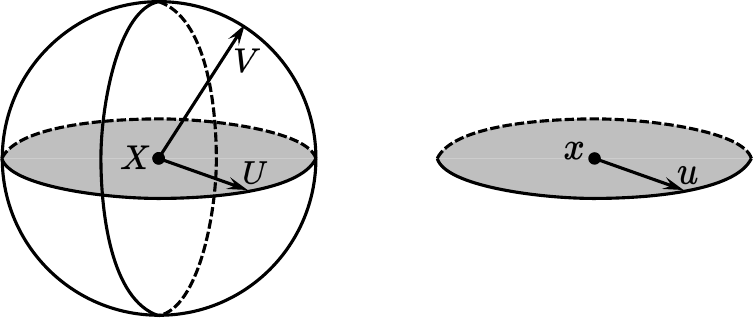}
    \caption{Illustration of our proposed SNS. In the HR space (left), randomly sampled noise $V$ is unlikely to find the exploitable space (grey). SNS overcomes this problem by first sampling noise $\bm{u}$ in the LR space (right) and then projecting it back to the HR $U$, which lies in the LR space.}
    \label{fig:demo:projection}
\end{figure}

Based on the above characterization, we propose to incorporate the weakness of the scaling stage into the noise sampling by Scaling-aware Noise Sampling (SNS). The high-level idea of SNS is illustrated in \Cref{fig:demo:projection}. Instead of sampling in the HR space $\SpaceHR$ (as is done by naively applying attacks to the full pipeline), we sample noise in the post-scaling space $\SpaceLR$ and project it back to $\SpaceHR$. This novel sampling strategy guides black-box attacks to search adversarial examples along the direction that hides the most perturbation through the scaling function. As a result, attacking the full pipeline $\ModelHR$ yields significantly better performance than attacking the standalone model $\ModelLR$.
However, projecting noise from $\SpaceLR$ to $\SpaceHR$ requires reversing the projection defined by $\Scale\circ\Defense$. This operation is not straightforward, especially when incorporating the pre-processor $\Defense$.

\subsubsection{Straightforward SNS}
\label{sec:method:sns:naive}

We first notice that reversing the projection of $\Scale\circ\Defense$ is an instance of the image-scaling attack~\cite{scaling2020}, as we also need to search for a point $U\in\SpaceHR$ that projects to the sampled $\bm{u}\in\SpaceLR$. As such, a straightforward solution is to solve the following image-scaling attack problem:
\begin{equation}
\label{eq:blackbox:projection}
    U^* \coloneqq\mathop{\arg\min}_{U\in\SpaceHR}\ \norm{(\Scale\circ\Defense)(X+U) - ((\Scale\circ\Defense)(X)+\bm{u})}_2^2,
\end{equation}
where $U^*\in\SpaceHR$ is the HR noise that lies on the LR space.

This approach is computationally prohibitive because we need to solve an optimization problem for every sampled noise. We overcome this problem with efficient SNS.

\subsubsection{Efficient SNS}
\label{sec:method:sns:efficient}
To improve the efficiency of SNS, we note that the final objective of SNS is to sample an HR \emph{noise} that lies in the LR space -- it is not necessary to find the \emph{exact} projection of a noise sample, as it is already random. Inspired by this observation, we find that an \emph{imprecise} projection suffices to guide the gradient estimation in black-box attacks. One imprecise yet efficient projection we find is the gradient of \Cref{eq:blackbox:projection}, written as
\begin{equation}\label{eq:method:sns}
    \tilde{U}\coloneqq\grad[U]{\norm{(\Scale\circ\Defense)(X+U) - ((\Scale\circ\Defense)(X)+\bm{u})}_2^2}.
\end{equation}

We use this more efficient solution to construct SNS, as summarized in \Cref{alg:sns}. It directs black-box attacks to efficiently search adversarial examples along the direction that hides the most perturbation through the scaling function.

\begin{algorithm}[tb]
	\caption{Scaling-aware Noise Sampling (SNS)}
	\begin{algorithmic}
		\Require Scaling procedure $\Scale\circ\Defense$, initial point $X\in\SpaceHR$.
		\Ensure A noise $U\in\SpaceHR$ that lies on the space $\SpaceLR$.
		\State Sample random noise $\bm{u}\in\SpaceLR$ (i.e., input space of $\ModelLR$).
		\State Compute $\tilde{U}\in\SpaceHR$ using \Cref{eq:method:sns}.
		\State Output $U\gets\tilde{U}$.
	\end{algorithmic}
	\label{alg:sns}
\end{algorithm}

\subsection{Incorporating Median Filtering Defenses}
\label{sec:method:median}
Although SNS is compatible with any projection defined by the scaling stage $\Scale\circ\Defense$, the defense $\Defense$ may have a side effect of hindering black-box attacks.
Our empirical evaluation in \Cref{sec:eval:preprocessing:median} reveals that \emph{median filtering defense can slow down the convergence of black-box attacks}.

We identify the root cause of this problem as the median function's robustness to outliers. When black-box attacks conduct line search along a fixed direction, the median filtering operation may not change its output in most of the searching steps. As a result, the attack converges slower and returns suboptimal results.
Without loss of generality, we illustrate how the HSJ attack~\cite{hopskip} conducts line search under the median function. Consider a starting point $\bm{x}=[1, 2, 3]$ and gradient $\bm{g}=[0, 1, 0]$. In this case, the line search procedure simply attempts $\s{\bm{x}+\bm{g}, \bm{x}+2\bm{g}, ...}$ until reaching the decision boundary. This procedure, however, only increases the perturbation without changing the output of the median function after reaching $\bm{x}+2\bm{g}$.

We overcome this problem by providing an estimate of the gradient that is more amenable to the line search. Since our attack estimates the gradient using noise sampled from \Cref{eq:method:sns}, we improve the gradient estimation by using a \emph{trimmed and weighted average} function in its backward pass. 
The formulation of our improved median function and the overall filtering defense can be found in \Cref{app:median}.

Our evaluation in \Cref{sec:eval:preprocessing:median} verifies that our improved estimation not only reduces the query number of searching adversarial examples but also exploits the scaling function as much as possible. This is different from other differentiable approximation approaches like BPDA~\cite{obfuscated}; e.g., using the identity function for approximation will not be able to exploit the scaling function.

\subsection{Incorporating Randomized Defenses}
\label{sec:method:random}

In this section, we explain how to modify SNS to incorporate randomized pre-processing defenses that protect the scaling function. Although such defenses can be easily circumvented in the white-box setting with expectation over transformation (EOT)~\cite{obfuscated,eot}, we note that \emph{directly applying EOT in black-box attacks is query-inefficient due to a large number of sampling operations}.

We overcome this challenge by computing the EOT \emph{offline} without querying the black-box model. That is, instead of attacking the \emph{expectation over the full pipeline}
\begin{equation}
	\EE_{\Defense\sim\mathcal{H}} \ModelHR(X) = \EE_{\Defense\sim\mathcal{H}} (\ModelLR\circ\Scale\circ\Defense)(X),
\end{equation}
we attack the \emph{expectation over pre-processors}
\begin{equation}\label{eq:expectation}
	(\ModelLR\circ\EE_{\Defense\sim\mathcal{H}} (\Scale\circ\Defense) )(X),
\end{equation}
where $\mathcal{H}$ is the space that draws a randomized defense $\Defense$. It is also possible to attack the expectation over the defense $\EE_{\Defense\sim\mathcal{H}}\Defense(X)$, but we found this expectation hard to compute, as image-scaling defenses $\Defense$ typically work jointly with the scaling function $\Scale$ to be effective.

The above strategy effectively reduces the number of samples in EOT to zero; this is possible because we are able to infer the knowledge of the scaling stage (see \Cref{sec:background:scaling-defenses:infer}). As such, we only need to change the original scaling stage $\Scale\circ\Defense$ in \Cref{eq:method:sns} to $\EE_{\Defense\sim\mathcal{H}} (\Scale\circ\Defense)$. In fact, we can derive the closed-form expectation for a particular defense, as we will discuss later in \Cref{sec:eval:preprocessing:random}.

\subsection{Plug-and-Play Integration with Black-box Attacks}
\label{sec:method:attack}
Since noise sampling is a critical step for most black-box attacks, our proposed SNS is directly applicable to all of these attacks, as illustrated in \Cref{fig:demo:attack}.
We demonstrate its generalizability with integrations of two representative black-box attacks: the boundary-based HSJ attack~\cite{hopskip} and the optimization-based Sign-OPT attack~\cite{signoptattack}. We refer to attacks on the model $\ModelLR$ as LR attacks, and our improved attacks on the full pipeline $\ModelHR$ as HR attacks.

\textbf{High-Resolution HSJ Attack.}
HSJ~\cite{hopskip} extends the boundary attack~\cite{boundaryattack} by walking near the decision boundary while adopting noise sampling to improve the gradient estimation. We apply SNS to guide its gradient estimation to hide perturbation. The detailed algorithm of our HR HSJ attack can be found in \Cref{app:integration:hsj}.

\textbf{High-Resolution Sign-OPT Attack.}
Sign-OPT~\cite{signoptattack} optimizes a direction for minimal distance to the decision boundary. We apply SNS to guide its gradient to search a direction that meanwhile causes minimal perturbation before scaling. The detailed algorithm of our HR Sign-OPT attack can be found in \Cref{app:integration:opt}.

\section{Evaluation}
\label{sec:eval}

Finally, we perform an empirical evaluation of our improved HR black-box attacks and five state-of-the-art defenses designed to protect the scaling procedure. Our evaluation is designed to answer the following questions.

\textbf{Q1: Can we improve black-box attacks by exploiting the scaling function to hide adversarial perturbation?}

We observe a significant improvement when LR black-box attacks leverage our proposed SNS to attack the entire ML pipeline. With the same query budget, our HR attacks generate adversarial examples with less perturbation. We demonstrate this problem in a real-world Image Analysis API with decision-based and transfer-based black-box attacks. 

\textbf{Q2: Can we still improve black-box attacks when the scaling function is protected by defenses?}

Our HR black-box attacks can still outperform their LR primitives under four out of five state-of-the-art defenses. These defenses include median filtering and all three detection defenses. We also analyze why these defenses fail to protect the scaling function despite their success in preventing standard image-scaling attacks.

\subsection{Evaluation Setup}
\label{sec:eval:setup}

\textbf{Dataset and Models.}
We use ImageNet~\cite{imagenet} and CelebA~\cite{celeba} datasets.
For ImageNet, we randomly choose 1,000 images whose scaling ratio is at least 3 and downscale them to $672\times672$; the target model is a pre-trained ResNet-50 model~\cite{resnet} that accepts input images of size $224\times224$.
For CelebA, we randomly choose 1,000 images and rescale their faces to $672\times672$; the target model is a pre-trained ResNet-34 model that accepts facial images of size $224\times224$ and predicts the \texttt{Mouth\_Slightly\_Open} attribute.
We provide more details of these datasets and models in \Cref{app:setup:models}.
We also include decision-based and transfer-based attacks on the Tencent Image Analysis API, whose details and settings can be found in \Cref{app:setup:api}.

\textbf{Attacks and Setup.}
We implement HR black-box attacks based on the HSJ and Sign-OPT attacks as described in \Cref{sec:method:attack}. 
We use OpenCV's linear scaling algorithm to represent the vulnerable scaling algorithm~\cite{scaling2020}.
We also provide evaluations of our HR attacks and the above defenses in the white-box setting in \Cref{app:whitebox}. Our code is available at \url{https://github.com/wi-pi/rethinking-image-scaling-attacks}.

\textbf{Evaluation Metrics.}
We use standard metrics:
(1) scaled \LL{2}-norm quantifies the adversarial perturbation divided by the scaling ratio to compare perturbation across different resolutions;
(2) attack success rate (ASR) at various scaled \LL{2}-norm thresholds under a particular query budget.

\begin{figure*}[tb]
	\centering
	\begin{subfigure}[t]{0.24\linewidth}
		\includegraphics[width=\linewidth]{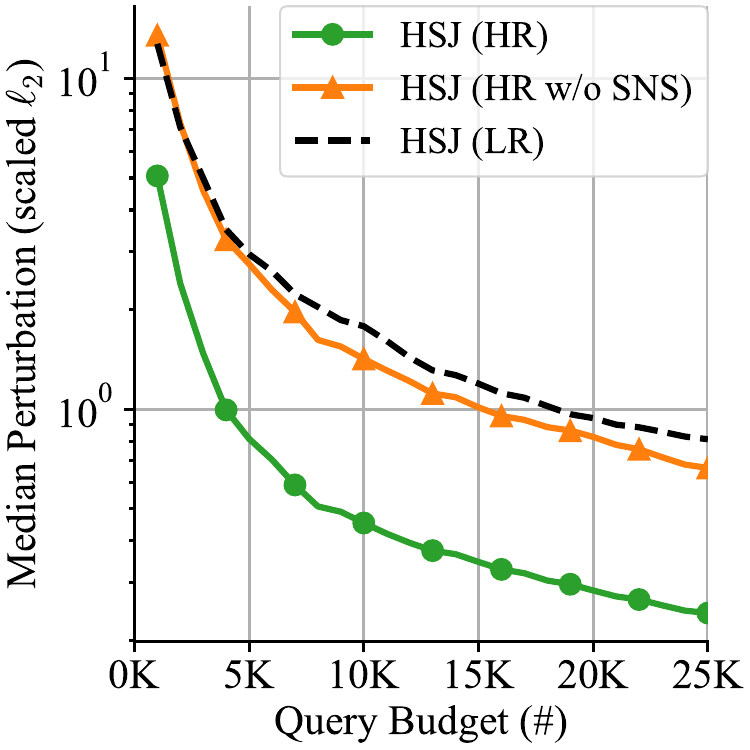}
		\caption{HSJ Attack}
	   \label{fig:eval:none:hsj-l2}
	\end{subfigure}
	\begin{subfigure}[t]{0.24\linewidth}
		\includegraphics[width=\linewidth]{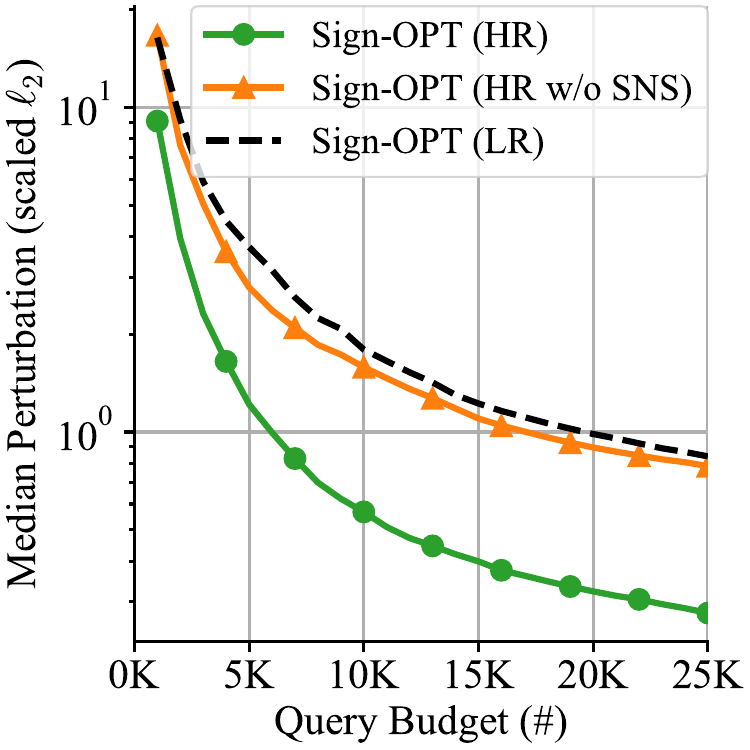}
		\caption{Sign-OPT Attack}
	   \label{fig:eval:none:opt-l2}
	\end{subfigure}
	\begin{subfigure}[t]{0.24\linewidth}
		\includegraphics[width=\linewidth]{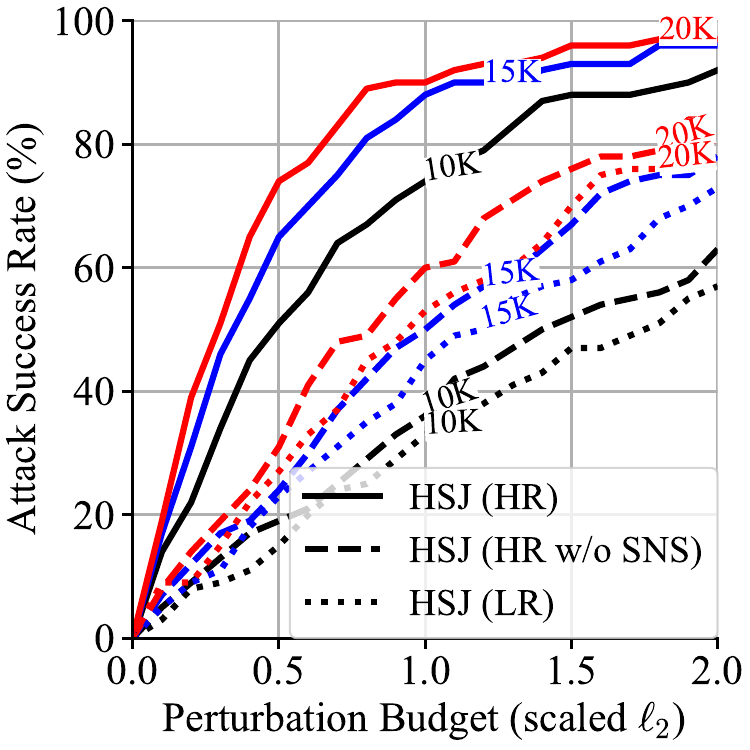}
		\caption{HSJ Attack}
	   \label{fig:eval:none:hsj-sar}
	\end{subfigure}
	\begin{subfigure}[t]{0.24\linewidth}
		\includegraphics[width=\linewidth]{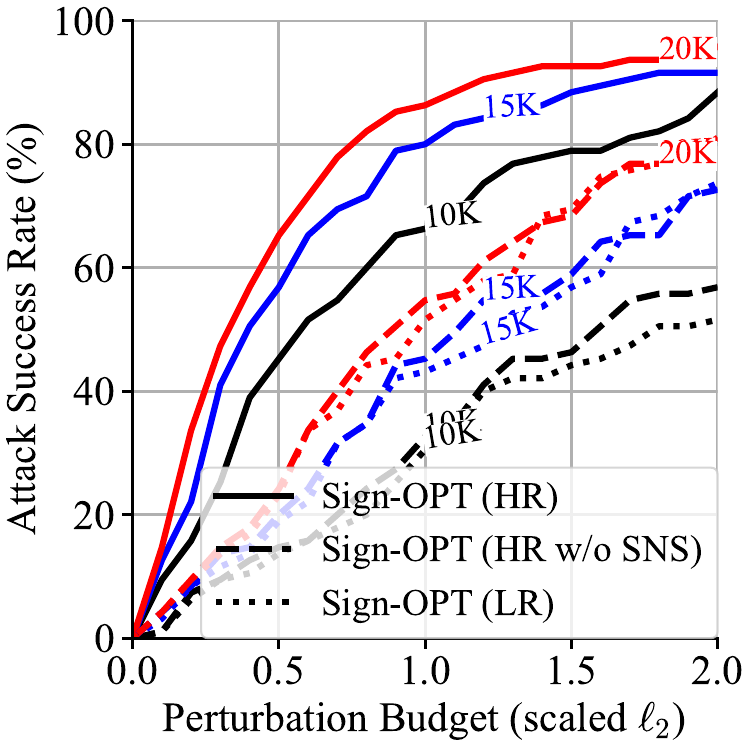}
		\caption{Sign-OPT Attack}
	   \label{fig:eval:none:opt-sar}
	\end{subfigure}
	\caption{Comparison of our HR HSJ and Sign-OPT attacks with their LR primitives \emph{under no defenses}. (a) and (b) compare the adversarial perturbation under different query budgets. (c) and (d) compare the attack success rate under different perturbation and query budgets. We include an ablation study to evaluate the effectiveness of our proposed SNS technique.}
	\label{fig:eval:none}
\end{figure*}

\subsection{Evaluation of Q1: Undefended Image Scaling}
\label{sec:eval:none}

In this experiment, we examine if black-box attacks can exploit the scaling function to improve their performance. For each query budget $q\in\s{1000, 2000, ..., 25000}$, we use the LR and HR attacks to generate a set of LR and HR adversarial examples, respectively. After that, we compare the perturbation between these two sets of adversarial examples.

\textbf{Evaluation of the Perturbation.} \Cref{fig:eval:none:hsj-l2,fig:eval:none:opt-l2} show the \emph{median perturbation} generated by attacks when given different query budgets. When there is no defense to protect the scaling function, our HR attacks can reduce the perturbation significantly.

\textbf{Evaluation of the Attack Success Rate.} \Cref{fig:eval:none:hsj-sar,fig:eval:none:opt-sar} show the \emph{attack success rate} of LR and HR attacks when given different perturbation budgets. When there is no defense to protect the scaling function, our HR attacks boost the success rate by a large margin. The solid lines of HR attacks are way above the dotted lines of LR attacks.

\textbf{Ablation Study (SNS).} We have included an ablation study in \Cref{fig:eval:none} by disabling our proposed SNS. HR attacks without SNS reduce to similar performance as LR attacks. It shows that simply attacking the entire pipeline cannot exploit the scaling function to gain benefits. We highlight some of the comparisons in \Cref{tab:highlight}.

We also provide an ablation study that compares the performance between the straightforward SNS and efficient SNS discussed in \Cref{sec:method:sns}, where we implement the straightforward SNS by solving \Cref{eq:blackbox:projection} using gradient descent with the Adam~\cite{adam} optimizer (1000 steps, 0.01 learning rate), which decreases the objective function to around 0.1. Due to the prohibitive computational cost, we only compare the attack effectiveness between precise and imprecise projections over 50 images. \Cref{tab:snseq12} shows that \Cref{eq:method:sns} loses little attack effectiveness while avoiding the cost of \Cref{eq:blackbox:projection}. It confirms our insight that improving the precision of a noise is not necessary --- as long as the noise lies in the desired subspace, estimating gradients using such noise suffices to incorporate the vulnerability.

\textbf{Ablation Study (Scaling Ratio).} We provide an ablation study in \Cref{tab:celeba} to examine the effectiveness of our chosen scaling ratio. Here, the attacker chooses a scaling ratio of 3 and 5. The results show that the attacker is free to choose a larger scaling ratio to yield even better results (under the assumption that the target model permits such large images).

\begin{figure}
    \centering
    \begin{subfigure}[t]{0.48\linewidth}
        \includegraphics[width=\linewidth]{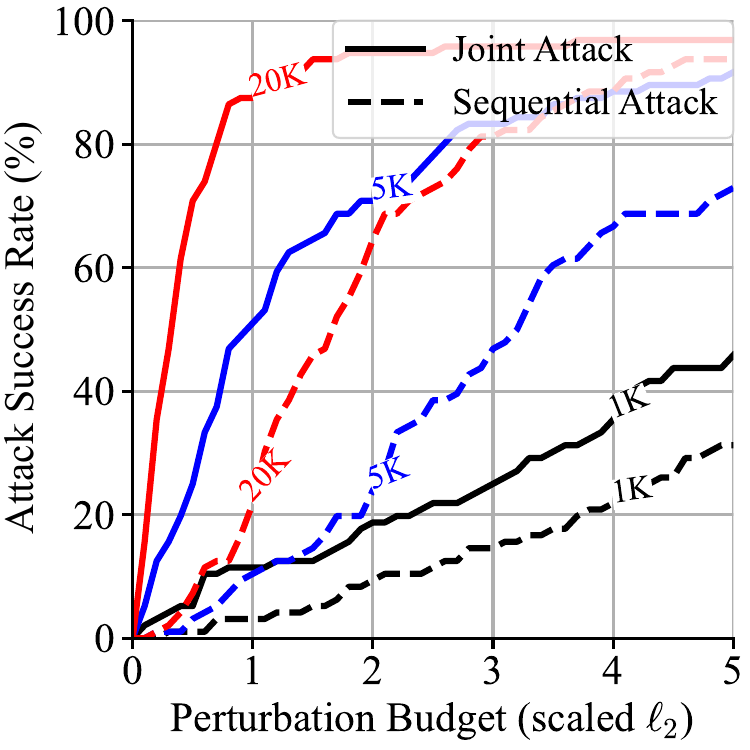}
        \caption{No Defense}
        \label{fig:exp:gen-vs-hide:1}
    \end{subfigure}
    \begin{subfigure}[t]{0.48\linewidth}
        \includegraphics[width=\linewidth]{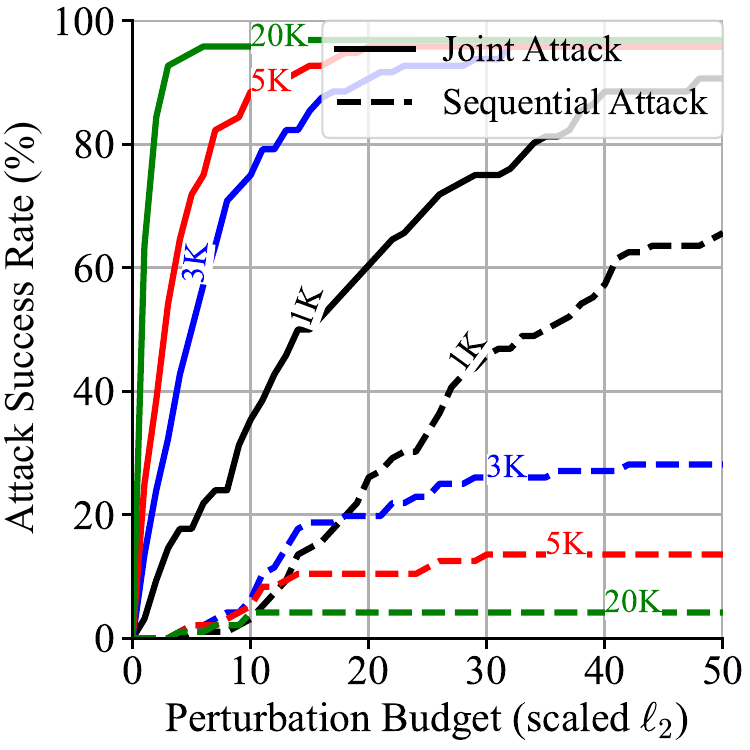}
        \caption{Median Filtering Defense}
        \label{fig:exp:gen-vs-hide:2}
    \end{subfigure}
    \caption{Comparison of attack strategies. Our joint attack is significantly better than the naive sequential combination of image-scaling attacks and adversarial examples.}
    \label{fig:exp:gen-vs-hide}
\end{figure}

\textbf{Ablation Study (Attack Strategy).}
We evaluate a \emph{sequential} attack (detailed in \Cref{app:sequential}) against the scaling function and ML model in \Cref{fig:exp:gen-vs-hide}; our joint attack outperforms the sequential attack by a large margin. Interestingly, the sequential attack in \Cref{fig:exp:gen-vs-hide:2} becomes worse when the target adversarial example was obtained with more queries, as opposed to the joint attack. It suggests that naively hiding a pre-generated adversarial example under the defense is harder than directly targeting the whole pipeline.

\begin{table}[t]
\centering
\caption{Comparison of HR attacks on CelebA with 10K queries using the straightforward and efficient SNS.}
\label{tab:snseq12}
\resizebox{\columnwidth}{!}{%
\begin{tabular}{@{}c|cccccc@{}}
\toprule
\multirow{2}{*}{Attacks}       & \multirow{2}{*}{\LL{2}} & \multicolumn{5}{c}{ASR under different \LL{2} budgets} \\
 &                         & 1.0     & 2.0    & 3.0    & 4.0    & 5.0     \\ \midrule
HSJ (HR) + Eq.\ (\ref{eq:blackbox:projection}) & \textbf{1.68}           & \textbf{22.9}\%  & \textbf{63.7}\% & \textbf{85.7}\% & \textbf{97.1}\% & \textbf{100.0}\% \\
HSJ (HR) + Eq.\ (\ref{eq:method:sns}) & 1.72                    & 20.0\%  & 60.0\% & 82.9\% & \textbf{97.1}\% & \textbf{100.0}\% \\ \bottomrule
\end{tabular}
}
\end{table}

\begin{table}[t]
\centering
\caption{Comparison of HR and LR attacks on CelebA with certain query budgets and image sizes. ASR is the attack success rate under perturbation budget $\ell_2=2.0$.}
\label{tab:celeba}
\resizebox{\columnwidth}{!}{%
\begin{tabular}{@{}l|cc|cc|cc@{}}
\toprule
\multicolumn{1}{c|}{\multirow{2}{*}{Attacks}} & \multicolumn{2}{c|}{Query = 10K} & \multicolumn{2}{c|}{Query = 15K} & \multicolumn{2}{c}{Query = 20K} \\
\multicolumn{1}{c|}{}                   & \LL{2}         & ASR             & \LL{2}         & ASR             & \LL{2}        & ASR             \\ \midrule
HSJ (LR, 224$\times$224)                       & 5.26           & 10.8\%          & 4.16           & 14.0\%          & 3.58          & 18.3\%          \\
HSJ (HR, 672$\times$672)                       & 1.73           & 59.6\%          & 1.37           & 74.5\%          & 1.17          & 86.2\%          \\
HSJ (HR, 1120$\times$1120)                     & \textbf{1.03}           & \textbf{90.4\%}          & \textbf{0.82}           & \textbf{98.9\%}          & \textbf{0.70}          & \textbf{100.0\%}          \\ \bottomrule
\end{tabular}
}
\end{table}

\begin{figure*}[htb]
	\centering
	\begin{subfigure}[t]{0.24\linewidth}
		\includegraphics[width=\linewidth]{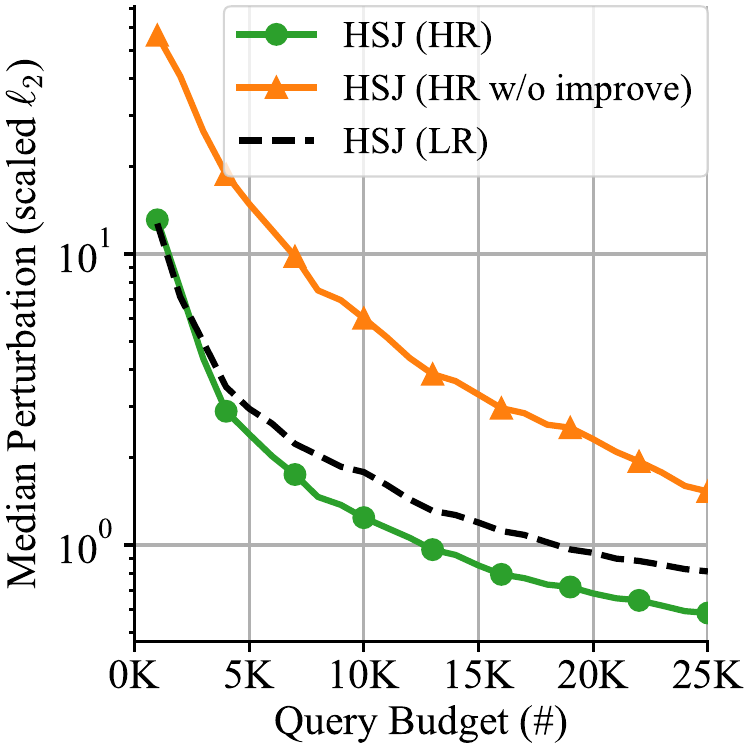}
		\caption{HSJ Attack}
		\label{fig:eval:median:hsj-l2}
	\end{subfigure}
	\begin{subfigure}[t]{0.24\linewidth}
		\includegraphics[width=\linewidth]{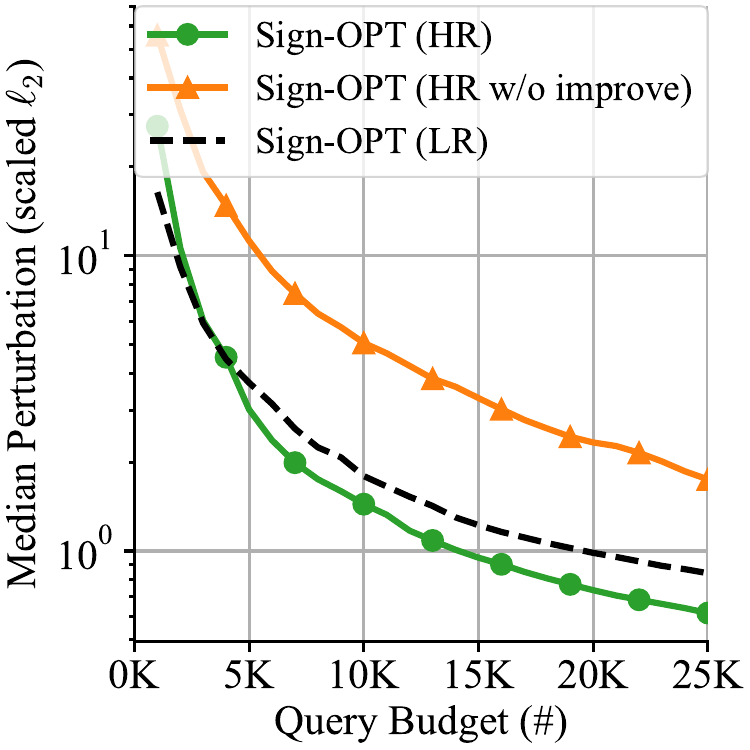}
		\caption{Sign-OPT Attack}
		\label{fig:eval:median:opt-l2}
	\end{subfigure}
	\begin{subfigure}[t]{0.24\linewidth}
		\includegraphics[width=\linewidth]{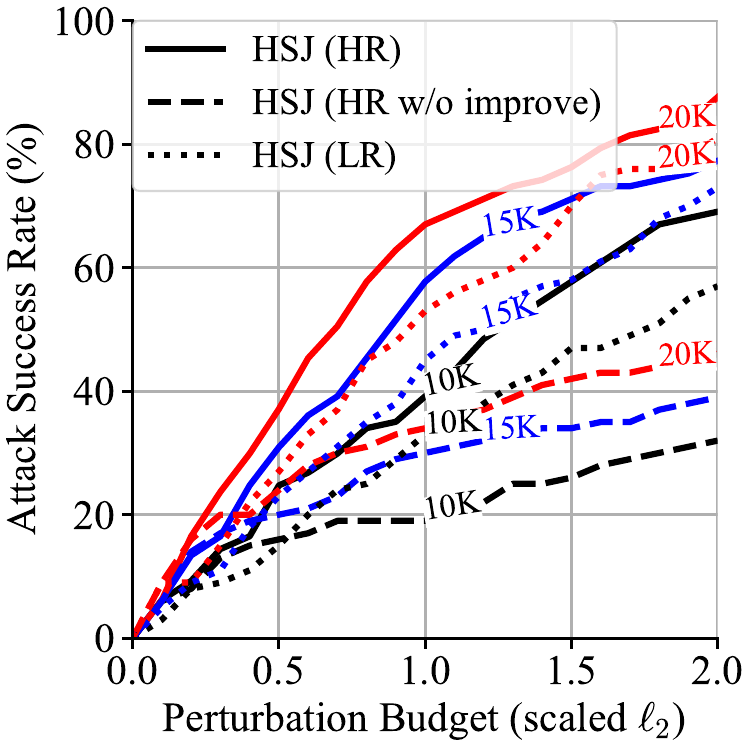}
		\caption{HSJ Attack}
		\label{fig:eval:median:hsj-sar}
	\end{subfigure}
	\begin{subfigure}[t]{0.24\linewidth}
		\includegraphics[width=\linewidth]{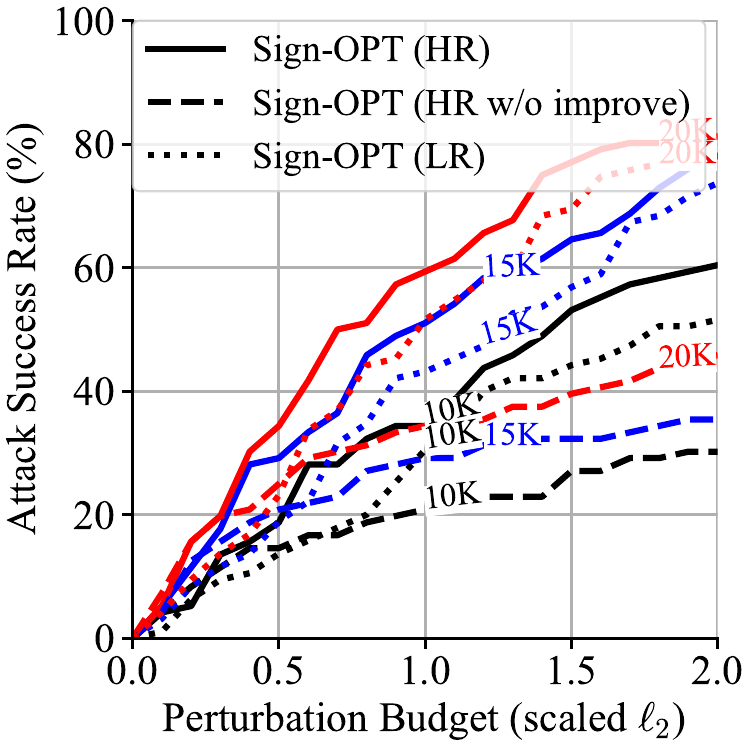}
		\caption{Sign-OPT Attack}
		\label{fig:eval:median:opt-sar}
	\end{subfigure}
	\caption{Comparison of our HR HSJ and Sign-OPT attacks with their LR primitives \emph{under the median filtering defense}. (a) and (b) compare the perturbation under different query budgets. (c) and (d) compare the attack success rate under different perturbation and query budgets. We include an ablation study that disables our improved gradient estimation for the median.}
	\label{fig:eval:median}
\end{figure*}

\subsection{Evaluation of Q2: Defended Image Scaling}
\label{sec:eval:preprocessing}

In this experiment, we examine if black-box attacks can still leverage the scaling function to improve their performance when there are defenses to protect the scaling function. We will discuss the details of each defense and analyze why they can (or cannot) prevent our improvement even if they have successfully blocked the image-scaling attack.

\subsubsection{Median Filtering}
\label{sec:eval:preprocessing:median}

\textbf{Defense Details.}
The median filtering defense sanitizes the input image by applying the median filter $\Kernel_\mathrm{med}$. To evade this defense, an adaptive attacker has to
perturb pixels in each window $\bm{w}$, such that the filtering output $w_m=\Kernel_\mathrm{med}(\bm{w})$ changes to the desired value $w_t$.
This defense was regarded as robust to an adaptive attacker that changes the filter's output by setting pixels within the range $R \coloneqq [w_m, w_t]$ to $w_t$. 
However, given that $\abs{R}\leq\abs{\bm{w}}/2$, the attacker needs only to modify at most half of the pixels to change the filtering output into a target value.

\textbf{Discussion.}
The above observation implies that median filtering's effectiveness relies on the (large) value of $\abs{R}$. In its original evaluation, $\abs{R}$ is always large as they only hide a non-adversarial image.
When considering adversarial perturbation, the target pixel $w_t$ will be close to the original output $w_m$, implying a small range $R=[w_m, w_t]$ that decreases the effectiveness.
Thus, although the median filtering defense can effectively prevent image-scaling attacks, it does not completely close the scaling function's vulnerability of stealthily hiding perturbation. 

\textbf{Evaluation.}
We evaluate the median filtering defense using our HR attacks and the improved gradient estimation described in \Cref{sec:method:median}. The comparison of HR and LR attacks are shown in \Cref{fig:eval:median}. As we can observe, our HR attacks still converge faster to a better solution than their LR primitives. As for attack success rate, HR attacks are able to outperform LR attacks even with 5K fewer queries. 

\textbf{Ablation Study.}
We have included an ablation study in \Cref{fig:eval:median} by disabling our improved gradient estimation of the median filtering defense. When our improved estimation is diabled, HR attacks are impractical and converge significantly slower. We have also highlighted some of the comparisons in \Cref{tab:highlight}.

\subsubsection{Randomized Filtering}
\label{sec:eval:preprocessing:random}

\textbf{Defense Details.}
The randomized filtering defense~\cite{scaling2020} sanitizes the input image by applying the random filter $\Kernel_\mathrm{rnd}$, which randomly picks a pixel from $\bm{w}$. To evade this defense, an adaptive attacker has to set every pixel in a window to the desired value. This defense was regarded as robust but lacked a rigorous argument.

\textbf{Discussion.}
We adopt the circumventing strategy in \Cref{sec:method:random} to analyze this defense. Specifically, we find that the expectation of randomized filtering scaling can be formulated as a uniform scaling procedure:
\begin{equation}\label{eq:random:expectation}
\EE_{\Defense\sim\mathcal{H}}(\Scale\circ\Defense)(X) = X\star \Kernel_\mathrm{u},
\end{equation}
where $\star$ denotes 2D convolution and $\Kernel_\mathrm{u}$ is the uniform scaling kernel (detailed in \Cref{app:random}). This observation shows that the randomized filter can process the input such that any followed scaling function will be made uniform in expectation, thereby leaving no space to hide perturbation stealthily. We discuss effective defenses and robust scaling algorithms with more details in \Cref{sec:background:scaling-defenses:robust,app:effective-defense}.

\textbf{Evaluation.}
We are not able to circumvent the randomized filtering defense to improve attacks. This result is verified in the white-box setting where we use PGD~\cite{pgd} to attack the entire pipeline (details in \Cref{app:whitebox:evaluate}). It shows that the randomized filtering defense can properly address the actual weakness of scaling functions.

\begin{table}[tb]
\centering
\caption{Comparison of HR and LR attacks with certain query budgets. ASR is the attack success rate under perturbation budget $\ell_2=1.0$. ($\ssymbol{1}$our SNS or improved gradient estimation is disabled, $\ssymbol{2}$under median filtering defense)}
\label{tab:highlight}
\resizebox{\columnwidth}{!}{%
\begin{tabular}{@{}l|cc|cc|cc@{}}
\toprule
\multicolumn{1}{c|}{\multirow{2}{*}{Attacks}} & \multicolumn{2}{c|}{Query = 10K} & \multicolumn{2}{c|}{Query = 15K} & \multicolumn{2}{c}{Query = 20K} \\
\multicolumn{1}{c|}{}                   & \LL{2}         & ASR             & \LL{2}         & ASR             & \LL{2}        & ASR             \\ \midrule[1pt]
HSJ (LR)                                & 1.78           & 33.0\%          & 1.19           & 45.0\%          & 0.94          & 53.0\%          \\ \midrule[0.2pt]
HSJ (HR$\ssymbol{1}$)                   & 1.42           & 36.0\%          & 1.01           & 50.0\%          & 0.82          & 60.0\%          \\
HSJ (HR)                                & \textbf{0.45}  & \textbf{74.0\%} & \textbf{0.35}  & \textbf{88.0\%} & \textbf{0.28} & \textbf{90.0\%} \\ \midrule[0.2pt]
HSJ (HR$\ssymbol{2}$$\ssymbol{1}$)      & 6.01           & 19.0\%          & 3.29           & 30.0\%          & 2.31          & 34.0\%          \\
HSJ (HR$\ssymbol{2}$)                   & \textbf{1.24}  & \textbf{39.2\%} & \textbf{0.85}  & \textbf{57.7\%} & \textbf{0.68} & \textbf{67.0\%} \\ \midrule[1pt]
Sign-OPT (LR)                           & 1.79           & 30.5\%          & 1.22           & 43.2\%          & 0.99          & 51.6\%          \\ \midrule[0.2pt]
Sign-OPT (HR$\ssymbol{1}$)              & 1.58           & 32.6\%          & 1.10           & 45.3\%          & 0.90          & 54.7\%          \\
Sign-OPT (HR)                           & \textbf{0.57}  & \textbf{66.3\%} & \textbf{0.40}  & \textbf{80.0\%} & \textbf{0.32} & \textbf{86.3\%} \\ \midrule[0.2pt]
Sign-OPT (HR$\ssymbol{2}$$\ssymbol{1}$) & 5.07           & 20.8\%          & 3.30           & 29.2\%          & 2.33          & 34.4\%          \\
Sign-OPT (HR$\ssymbol{2}$)              & \textbf{1.44}  & \textbf{34.4\%} & \textbf{0.95}  & \textbf{51.0\%} & \textbf{0.74} & \textbf{59.4\%} \\ \bottomrule
\end{tabular}
}
\end{table}

\begin{figure}[htb]
    \centering
    \begin{subfigure}[t]{0.8\linewidth}
        \includegraphics[width=\linewidth]{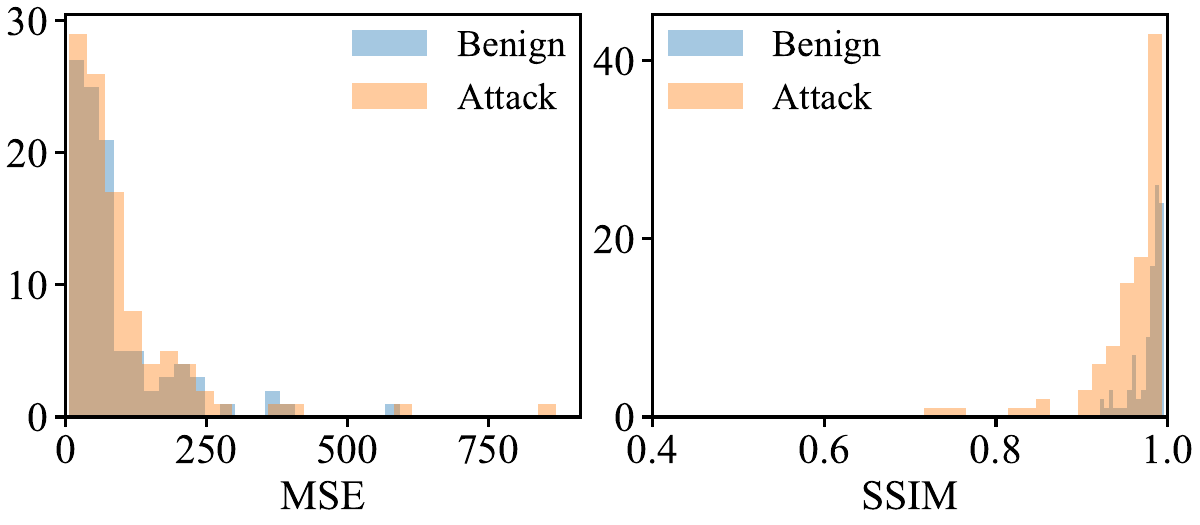}
        \caption{Undefended Scaling}
    \end{subfigure}
    \begin{subfigure}[t]{0.8\linewidth}
        \includegraphics[width=\linewidth]{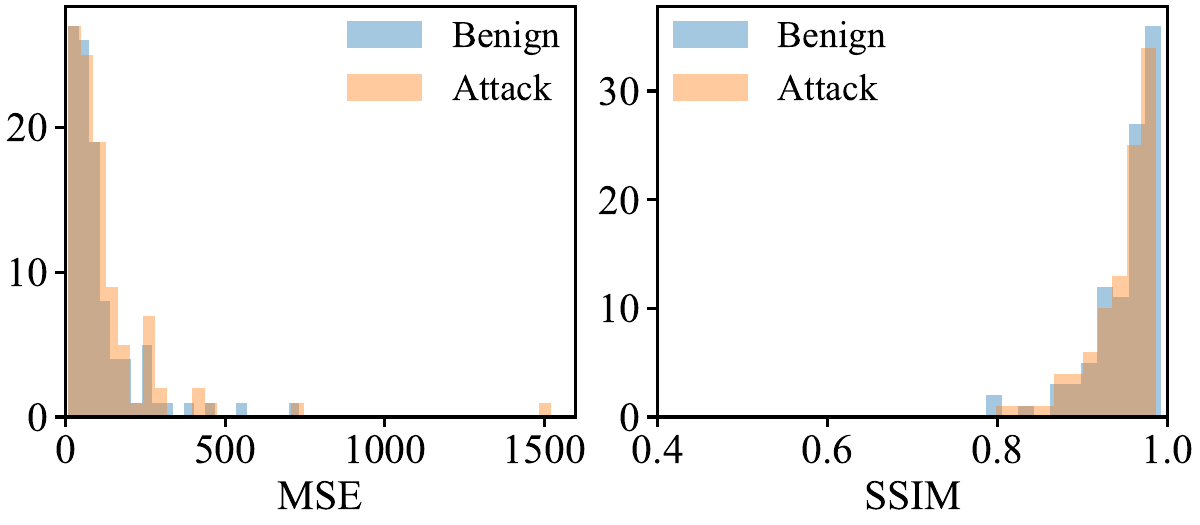}
        \caption{Median-defended Scaling}
        \label{fig:exp:detection:unscaling:b}
    \end{subfigure}
    \caption{The histogram of distortions as measured by the unscaling defense. Benign images and adversarial examples produced by our HR black-box attacks are indistinguishable.}
    \label{fig:exp:detection:unscaling}
\end{figure}

\begin{figure}[htb]
    \centering
    \includegraphics[width=0.24\linewidth]{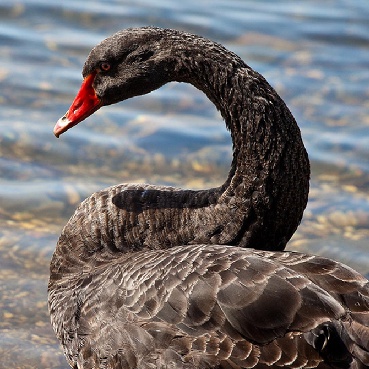}
    \hfill
    \includegraphics[width=0.24\linewidth]{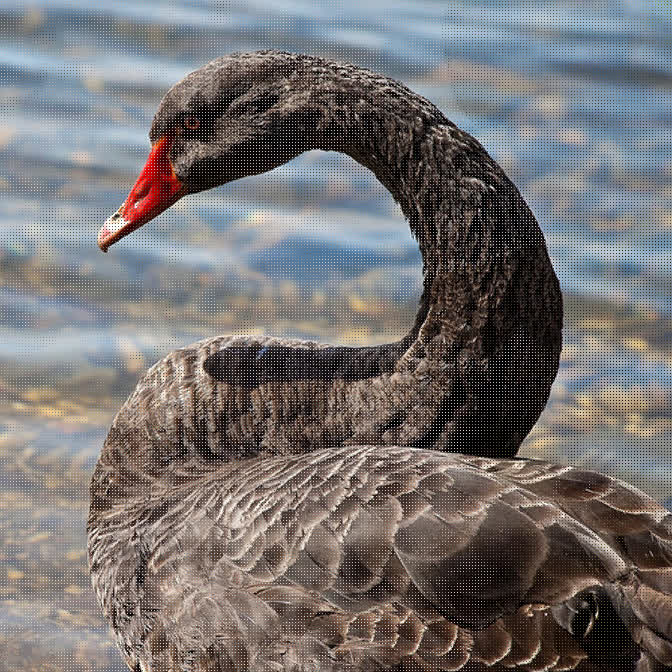}
    \hfill
    \includegraphics[width=0.24\linewidth]{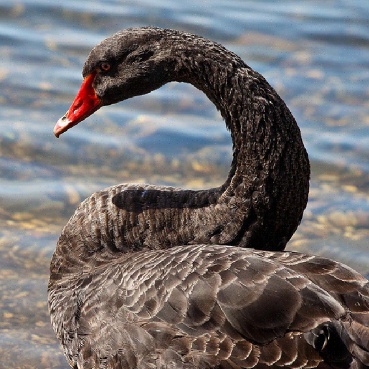}
    \hfill
    \includegraphics[width=0.24\linewidth]{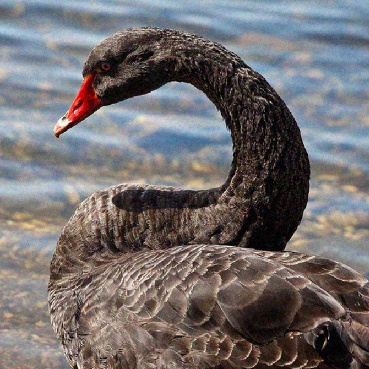}

    \vspace{0.3em}

    \begin{subfigure}[t]{0.24\linewidth}
        \includegraphics[width=\linewidth]{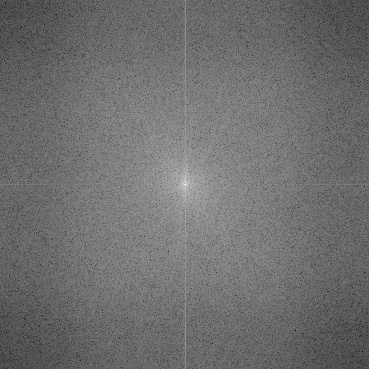}
        \caption{Benign}
    \end{subfigure}
    \hfill
    \begin{subfigure}[t]{0.24\linewidth}
        \includegraphics[width=\linewidth]{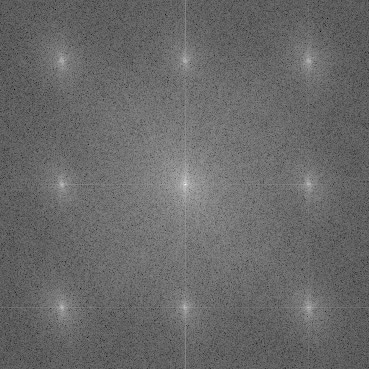}
        \caption{\centering Image- Scaling Attack}
        \label{fig:exp:detection:spectrum:bad}
    \end{subfigure}
    \hfill
    \begin{subfigure}[t]{0.24\linewidth}
        \includegraphics[width=\linewidth]{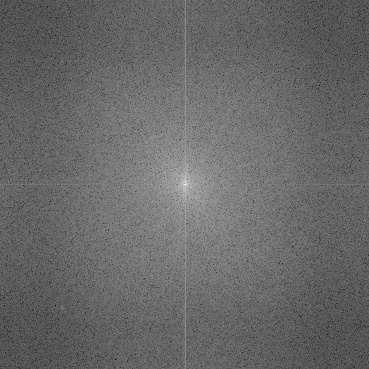}
        \caption{\centering HR-HSJ (undefended)}
    \end{subfigure}
    \hfill
    \begin{subfigure}[t]{0.24\linewidth}
        \includegraphics[width=\linewidth]{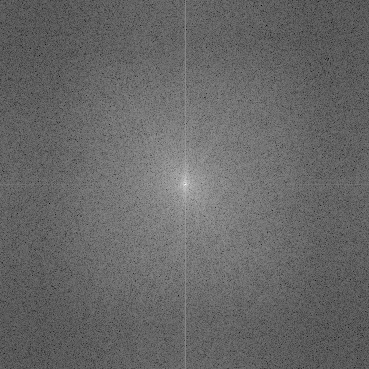}
        \caption{\centering HR-HSJ (median)}
        \label{fig:exp:detection:spectrum:median}
    \end{subfigure}
    \caption{The centered spectrum of benign and adversarial examples. Our HR attacks do not exhibit artifacts like (b).}
    \label{fig:exp:detection:spectrum}
\end{figure}

\subsubsection{Spatial Domain Detection}
\label{sec:eval:detection}

In the spatial domain, \citet{detection2020} leverage \emph{unscaling} and \emph{min-filtering} to reveal the injected perturbation through processing the input image $X$ with some function $T$. 
For example, the unscaling defense considers $T$ as the composition of downscaling and upscaling, which explicitly reveals the hidden perturbation. 
After that, they quantify the resulting distortion with perceptual metrics like MSE and SSIM. One could model the distortion score as $t(X) \coloneqq \mathrm{MSE}(X - T(X))$ and employ a threshold-based detector to determine whether the input image $X$ is benign or perturbed.
Our evaluation in \Cref{fig:exp:detection:unscaling} shows that these defenses cannot detect the hidden perturbation when they are sufficiently small. Results from the minimum-filtering defense are not shown as they show similar observations.

\subsubsection{Frequency Domain Detection}
In the frequency domain, \citet{detection2020} examine the number of peaks in the spectrum image, as it assumes injected perturbations manifest as high-frequency and high-energy noise. It applies a low-pass filter (with a predefined threshold) on the input's spectrum image to reveal such peaks. However, our HR attacks in \Cref{fig:exp:detection:spectrum} do not have such artifacts. A sophisticated defender could employ learning-based detectors to detect the hidden perturbation; we leave its black-box circumvention to future work.

\subsection{Attacking Cloud API}
\label{sec:eval:api}
\begin{figure}[t]
	\centering
    \begin{subfigure}[t]{0.48\linewidth}
        \includegraphics[width=\linewidth]{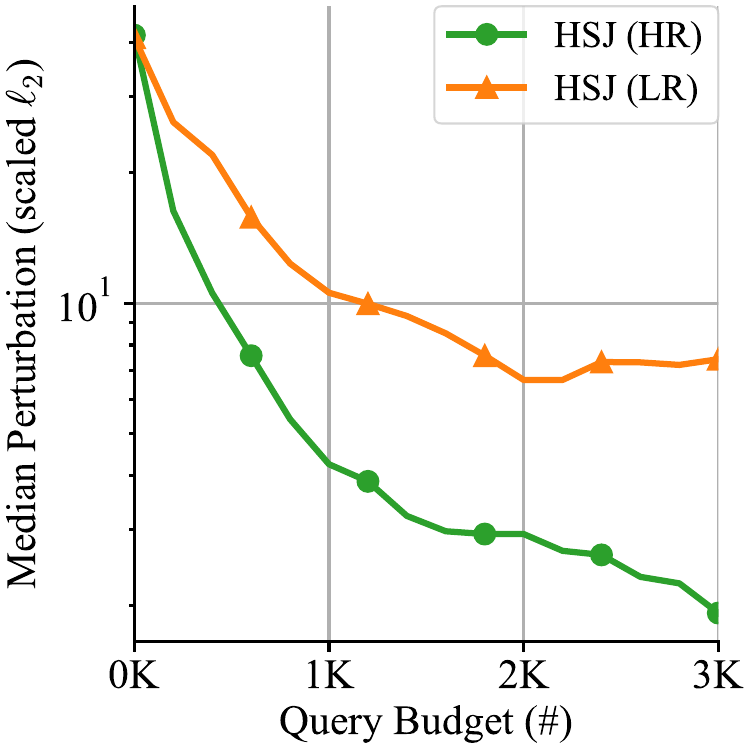}
        \caption{Perturbation (scaled $\ell_2$)}
        \label{fig:exp:api:1}
    \end{subfigure}
    \begin{subfigure}[t]{0.48\linewidth}
        \includegraphics[width=\linewidth]{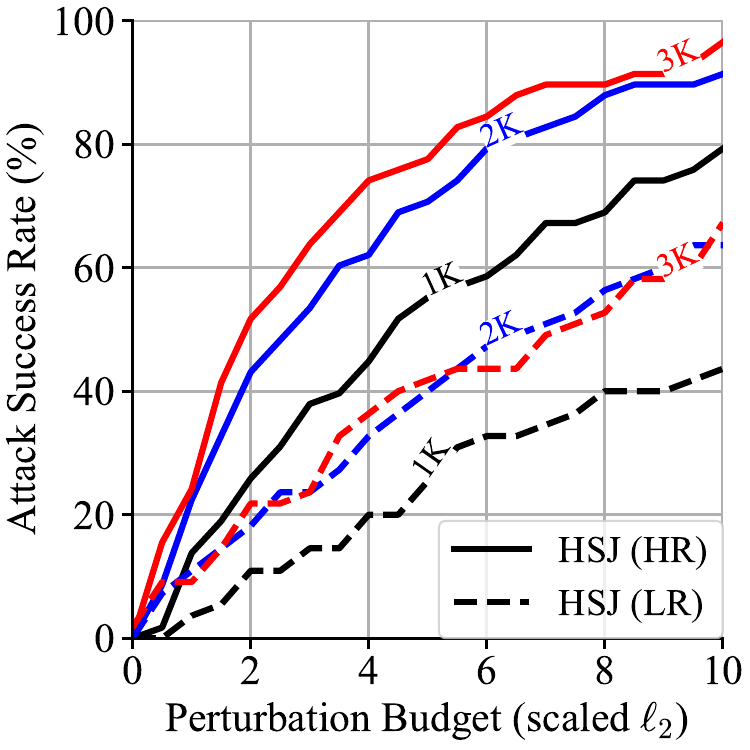}
        \caption{Attack Success Rate (\%)}
        \label{fig:exp:api:2}
    \end{subfigure}
    \caption{Comparison of the decision-based LR and HR HSJ attacks on the Tencent Cloud API.}	
    \label{fig:exp:api-hsj}
\end{figure}

Finally, we conduct black-box attacks on the Tencent Cloud API. This experiment demonstrates that the attacker can also exploit the scaling function in online APIs to improve their attack. 
For decision-based online attacks, we test 100 ImageNet images, each with 3K queries (\$1.18 USD per image).
The results are shown in \Cref{fig:exp:api-hsj} and highlighted in \Cref{tab:2}, where the HR attack significantly outperforms its LR counterpart.

\begin{figure}[t]
	\centering
    \begin{subfigure}[t]{0.48\linewidth}
        \includegraphics[width=\linewidth]{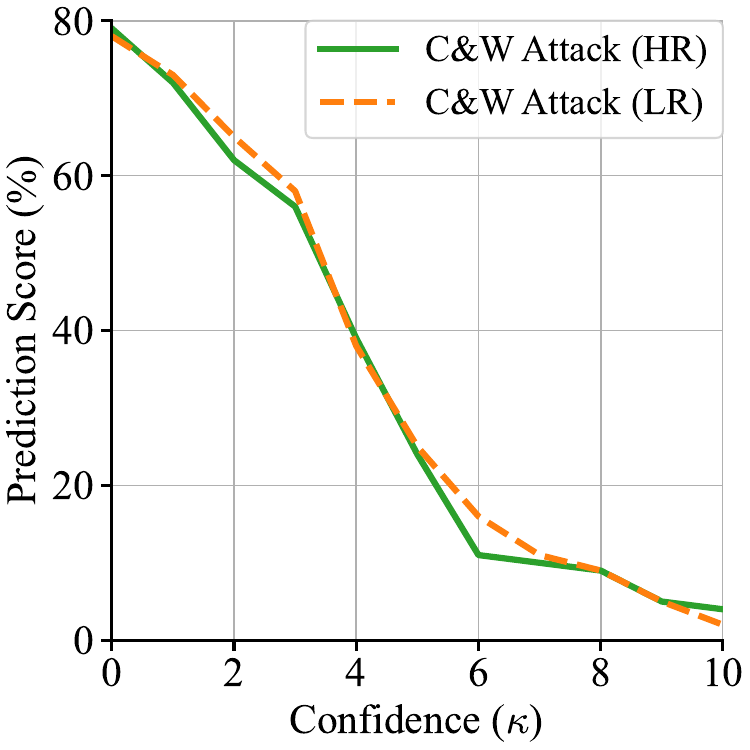}
        \caption{Prediction Score (\%)}
        \label{fig:exp:api-cw:1}
    \end{subfigure}
    \begin{subfigure}[t]{0.48\linewidth}
        \includegraphics[width=\linewidth]{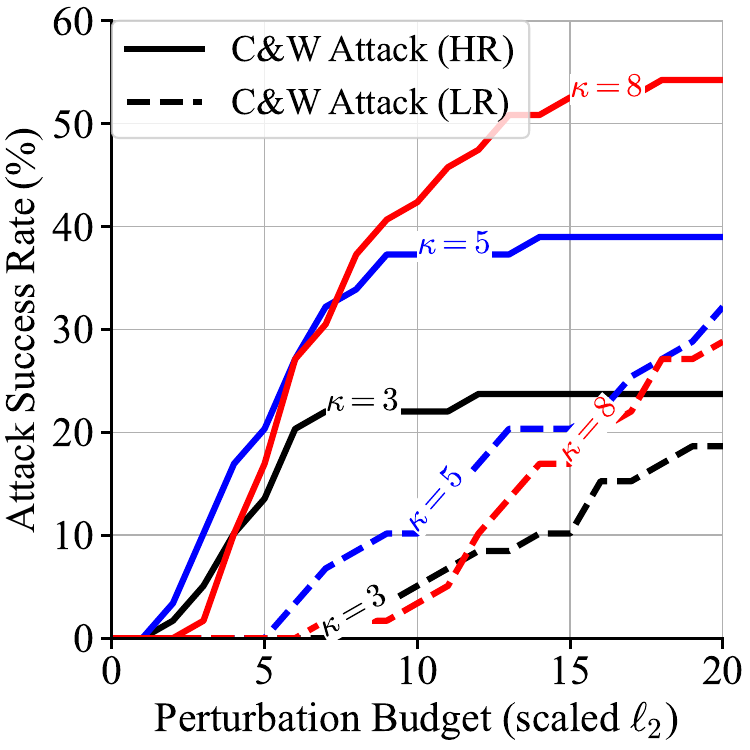}
        \caption{Attack Success Rate (\%)}
        \label{fig:exp:api-cw:2}
    \end{subfigure}
    \caption{Comparison of the transfer-based LR and HR C\&W attacks on the Tencent Cloud API. The reported prediction score is returned by the API and indicates how confident it predicts the input as its ground-truth label.}	
    \label{fig:exp:api-cw}
\end{figure}

The transfer-based attacks in \Cref{fig:exp:api-cw} confirm that C\&W attack~\cite{carlini2019evaluating} can achieve significantly higher transferability and success rates than their LR counterparts. In particular, \Cref{fig:exp:api-cw:1} show that HR attacks do not misuse the confidence parameter to overclaim improvements.

\begin{table}[t]
\centering
\caption{Highlighted comparison of the decision-based LR and HR HSJ attacks on Tencent Cloud API. ASR is the attack success rate under perturbation budget $\ell_2=2.0$.}
\label{tab:2}
\resizebox{0.9\columnwidth}{!}{%
\begin{tabular}{@{}lcccccc@{}}
\toprule
\multicolumn{1}{c}{\multirow{2}{*}{Attacks}} & \multicolumn{2}{c}{Query = 1K} & \multicolumn{2}{c}{Query = 2K} & \multicolumn{2}{c}{Query = 3K} \\
\multicolumn{1}{c}{}  & \LL{2}         & ASR              & \LL{2}         & ASR              & \LL{2}         & ASR              \\ \midrule
HSJ (LR)     & 10.57          & 10.9\%           & 6.64           & 18.2\%           & 7.42           & 21.8\%           \\
HSJ (HR)     & \textbf{4.24}  & \textbf{25.9\%}  & \textbf{2.93}  & \textbf{43.1\%}  & \textbf{1.92}  & \textbf{51.7\%}  \\ \bottomrule
\end{tabular}
}
\end{table}

\section{Discussion}

The evaluation of the trustworthiness of ML systems must consider the \emph{interplay} of different vulnerabilities. When weaknesses coexist in the different stages of the ML pipeline, defenders should carefully analyze if these weaknesses could amplify attacks designed to exploit any of them. This work shows that black-box attacks can be made stronger given the knowledge of a vulnerable preprocessing stage, scaling in our case. Further, it shows that defenses, which are effective against the standalone image-scaling attack, can be exploited to make the black-box attack stronger.

To prevent a false sense of security, the defender should also address the \emph{weakness} exploited by the attack rather than the attack itself. Although existing image-scaling defenses can successfully block the image-scaling attack, most of them fail to mitigate the underlying vulnerability of scaling algorithms properly. This leaves an open space for an attacker to jointly exploit other weaknesses in ML, like the adversarial example, as is done in this work.

\textbf{Limitations.}
In this paper, we mainly focus on the black-box attacks that estimate gradients and optimize for \LL{2} norm. There are other attacks that do not directly estimate gradients and optimize for the \LL{\infty} norm, such as RayS~\cite{rays}. In such cases, it is possible to integrate them by projecting their search direction to the exploitable subspace; for instance, adapting our \Cref{eq:method:sns} to optimize the direction instead of the noise. Despite, the benefits would still manifest in \LL{2} norm; the vulnerability of scaling algorithms mainly improve perceptual quality, so \LL{\infty} norm is inapplicable. Benefits in terms of \LL{\infty} norm would require future work to explore other vulnerabilities that discard a large amount of information in the magnitude of pixel values.

\section{Conclusion}
This paper explores the interplay between vulnerabilities of image scaling and ML models in the black-box setting. We propose a novel sampling strategy to make black-box attacks exploit the weakness of scaling functions. With our novel circumvention strategy, we show that 4 out of 5 state-of-the-art defenses, designed to protect the scaling stage, retain weaknesses that enable stronger black-box attacks.
The purpose of this work is to raise the concern of threats that jointly exploit different vulnerabilities, whereas current efforts focusing on defending against each vulnerability separately. Further work is necessary to identify and mitigate other threats that jointly target different ML components.

\section*{Acknowledgement}
We thank all anonymous reviewers for their insightful comments and feedback. This work is partially supported by the DARPA GARD program under agreement number 885000 and the NSF through awards: CNS-1838733, CNS1942014, and CNS-2003129.

\clearpage
\bibliography{main}
\bibliographystyle{icml2022}

\clearpage
\appendix
\FloatBarrier
\twocolumn[
\icmltitle{Supplementary Materials:  \\
The Interplay Between Vulnerabilities in Machine Learning Systems}
]

\icmltitlerunning{Supplementary Material: The Interplay Between Vulnerabilities in Machine Learning Systems}

\section{Background of Image Scaling}
In this section, we provide additional details of the image-scaling attack and the formulation of scaling algorithms.

\subsection{Image Scaling}
\label{app:scaling}
The scaling procedure $\Scale(\cdot)$ resizes a high-resolution (HR) source image $X\in\SpaceHR\coloneqq[0,1]^{m\times n}$ to the low-resolution (LR) output image $\bm{x}\in\SpaceLR\coloneqq[0,1]^{p\times q}$. The overall scaling ratio is defined as $\beta=\min\s{\beta_h, \beta_v}$, where $\beta_h=n/q$ and $\beta_v=m/p$ are the scaling ratios in two directions. In this paper, we only consider downscaling where $\beta>1$.

The scaling function can be implemented in different ways; we review two formulations that facilitate our analysis. Both formulations indicate that the standard scaling function is a linear operation, thus the post-scaling space $\SpaceLR$ can be viewed as a subspace of the pre-scaling space $\SpaceHR$.

\paragraph{Matrix Multiplication.}
\citet{scaling2019} conduct an empirical analysis of common scaling functions. They represent image scaling as matrix multiplications:
\begin{equation}\label{eq:scaling:matrix}
\bm{x} = \Scale(X) = L \times X \times R,
\end{equation}
where $L\in\RR^{p\times m}$ and $R\in\RR^{n\times q}$ are the two constant-coefficient matrices determined by the applied scaling function. They also provide an efficient strategy to deduce approximations of these matrices from given implementations.

\noindent\textbf{Convolution.}
\citet{scaling2020} interpret scaling as a convolution\footnote{More precisely, this should be cross-correlation~\cite{convolution-arithmetic}. But we will use the term convolution for consistency.} between the source image $X$ and a fixed linear kernel $\Kernel$ determined by the scaling algorithm:
\begin{equation}\label{eq:scaling:conv}
\bm{x} = \Scale(X) = X \star \Kernel,
\end{equation}
where $\star$ denotes the 2D convolution with proper padding and stride size to match the desired output shape.

\subsection{Image-Scaling Attacks}
\label{app:scaling-attacks}

\begin{figure}[tb]
    \centering
    \includegraphics[width=\linewidth]{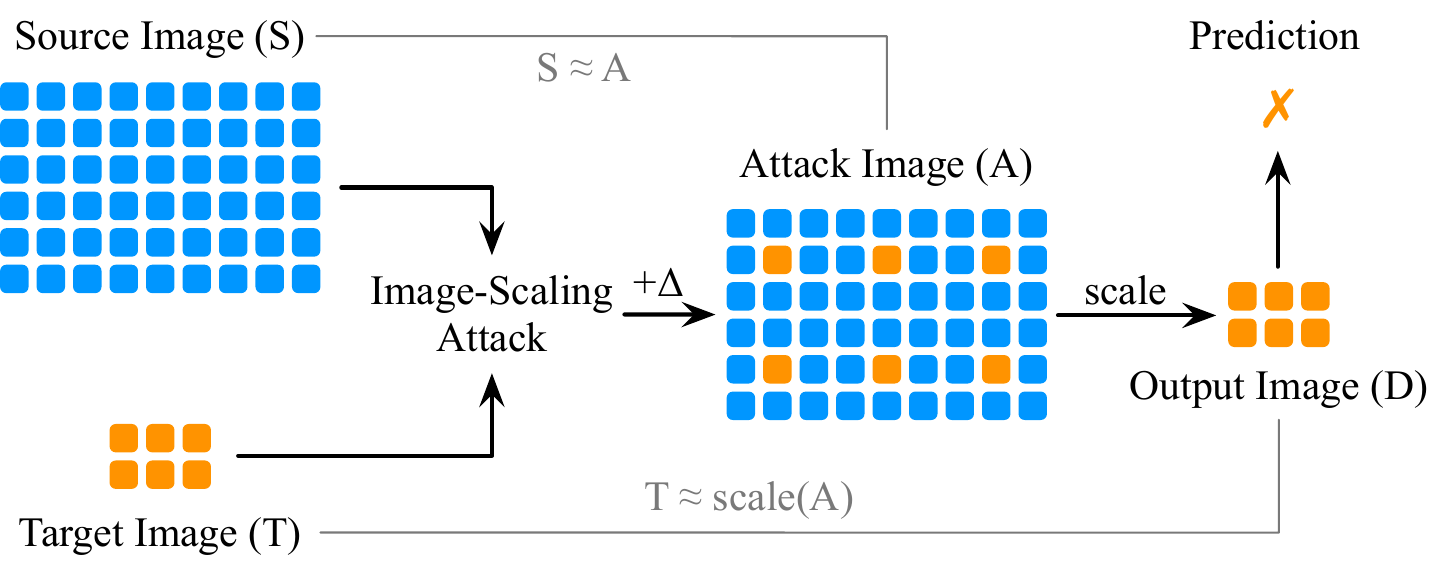}
    \caption{Illustration of the image-scaling attack~\cite{scaling2020}. The HR adversarial example $A$ looks similar to the clean image $S$, but changes into a target image $T$ after scaling. The color indicates pixels from different images.}
    \label{fig:demo:scaling-attack}
\end{figure}

The scaling attacks~\cite{scaling2019,scaling2020} target \emph{only} the scaling procedure in an ML system pipeline. They demonstrate that an attacker can exploit the scaling procedure to compromise an arbitrary downstream ML model. \Cref{fig:demo:scaling-attack} illustrates the pipeline of this attack. An adversary computes the attack image $A$ by adding imperceptible perturbations $\Delta$ to the source image $S$, such that it becomes similar to a target image $T$ (with a different label from $S$) after scaling, thereby fooling the downstream classifier.
They formulate the attack as a quadratic optimization problem:
\begin{equation}\label{eq:scaling-attack}
\min\ \norm{\Delta}_2^2 \quad \mathrm{s.t.}\ \norm{\Scale(S+\Delta)-T}_\infty\leq\epsilon,
\end{equation}
where the attack image $A\coloneqq S+\Delta$ satisfies the box constraint $A\in\br{0, 1}^{m\times n}$. 

Besides empirical attacks, \citet{scaling2020} conduct an in-depth analysis of common scaling algorithms and corresponding convolution kernels used in \Cref{eq:scaling:conv}. They identify the use of non-uniform kernels of a fixed width as the root cause for scaling attacks. Such kernels assign higher weights to a small set of \emph{vulnerable pixels} in the source image. For example, in \Cref{fig:demo:scaling-attack}, the attacker only needs to modify a few vulnerable (orange) pixels in the source image to change the scaling output completely.

The scaling attack works under the black-box setting. It only needs hundreds of decision-only queries to deduce the fixed scaling algorithm in an ML system~\cite{scaling2019}.

\begin{table*}[tb]
\small
    \centering
    \begin{tabular}{lcl}
        \toprule
        \textbf{Defense} & \textbf{Type} & \textbf{Technique} \\
        \midrule
        Median~\cite{scaling2020} & Pre-processing & Apply median filtering in each window to remove injected perturbation.\\
        Randomized~\cite{scaling2020} & Pre-processing & Randomly sample pixels in each window to remove injected perturbation.\\
        \midrule
        Unscaling~\cite{detection2020} & Detection & Down-scale and then up-scale an image to reveal and detect perturbation.\\
        Min-filtering~\cite{detection2020} & Detection & Apply minimum filtering to reveal injected perturbation.\\
        Spectrum~\cite{detection2020} & Detection & Identify more-than-one peaks in the centered spectrum.\\
        \bottomrule
    \end{tabular}
    \caption{Techniques used by recent scaling defenses.}
    \label{fig:scaling-defenses}
\end{table*}

\subsection{Image-Scaling Defenses}
\label{app:scaling-defense}

Researchers proposed several add-on defenses against the scaling attack; these defenses fall into two categories: prevention and detection defenses. We review five state-of-the-art defenses as summarized in \Cref{fig:scaling-defenses}.

\subsubsection{Prevention Defenses}
\label{sec:background:scaling-defenses:prevention}

\citet{scaling2020} propose the only two prevention defenses, i.e., median and randomized filtering. Both defenses apply filtering operations to sanitize the input image before scaling. Specifically, they reconstruct each vulnerable pixel by a median or a randomly picked pixel within a sliding window. As a result, the attacker has to perturb a significantly larger number of pixels within a window to evade these defenses.
Finally, they claim that the median filtering defense is more practical; the randomized filtering defense could hurt the downstream classifier's performance. We analyzed these defenses in more detail in \Cref{sec:eval:preprocessing}.

\subsubsection{Detection Defenses}
\label{sec:background:scaling-defenses:detection}

\citet{detection2020} propose three detection defenses using spatial and frequency transformations: unscaling, minimum-filtering, and centered spectrum. These transformations result in discernible differences when applied to benign and attack images.
The unscaling invokes downscale and upscale operations sequentially to reveal the injected image. If the input image is benign, this procedure should reveal a similar image to the original input one. In case of an attack image, this sequence would reveal a different image. Similarly, minimum filtering reveals such differences using the minimum filter operation. By measuring this difference, they construct a threshold-based detector using mean squared error (MSE) and structural similarity index (SSIM).
They also notice that the attack perturbation manifests as high-energy and high-frequency noise, which is detectable by examining the spectrum image. We analyzed these defenses in more detail in \Cref{sec:eval:preprocessing}.

\subsubsection{Infer the Knowledge of Defenses}
\label{sec:background:scaling-defenses:infer}
One can easily infer the knowledge of deployed scaling function and defense even in the black-box setting.
Existing image-scaling attacks~\cite{scaling2019} brute-force the scaling function $\Scale$ and the network's input space $\SpaceLR$ with hundreds of black-box queries. They run image-scaling attacks with all combinations of standard scaling functions and input sizes until succeed. Note that this inferred knowledge can be reused in the following attacks, as this setting is typically fixed for a deployed ML model. Finally, one can easily extend this method to infer the knowledge of the defense $\Defense$, as the number of defense parameters is also limited.

\subsubsection{Robust Scaling Algorithms}
\label{sec:background:scaling-defenses:robust}
Besides the above add-on defenses, \citet{scaling2020} identify several scaling algorithms that are naturally robust to the scaling attack. These algorithms are robust because they use either uniform kernels or dynamic kernel widths. For instance, the area (i.e., uniform) scaling algorithm convolves the input image (of scaling ratio $\beta$) with a uniform kernel $\Kernel_\mathrm{u}$ of size $\beta\times\beta$, where each entry of $\Kernel_\mathrm{u}$ is set to $1/\beta^2$ -- the kernel considers each pixel in the window equally. Thus, the attacker cannot find vulnerable pixels to inject another image stealthily.

Ideally, such algorithms should be part of the ML pipeline, but the default scaling algorithm in common ML frameworks is not robust~\cite{scaling2020}. Thus, switching to a different (robust) algorithm faces compatibility issues, such as changing dependent libraries, performance degradation, and even model retraining. As such, deployed ML systems would prefer add-on defenses that can easily fit as plugin modules~\cite{detection2020}. However, our discussion and evaluation in \Cref{sec:eval:preprocessing} show that ML systems need to avoid such add-on defenses; they should deploy scaling algorithms that are robust by design.

\subsubsection{Effective Image-Scaling Defenses}
\label{app:effective-defense}
We discuss what it means for a scaling defense to be effective (or ``robust'' under the terminology from \citet{scaling2020}). Prevention defenses are effective if images do not change their appearance after the defended scaling, such that the attacker cannot hide a large perturbation stealthily. Since a defended scaling procedure $\Scale\circ\Defense$ can be viewed as a new scaling function, we note that it must also satisfy the argument from \cite{scaling2020} about robust scaling algorithms (refer \Cref{sec:background:scaling-defenses:robust}). This simple observation indicates that \textbf{an effective prevention defense should process the input, such that the followed scaling function can weight all pixels uniformly}.
As for detection defenses, they are effective if they can detect the attack with acceptable false acceptance and rejection rates.

\section{Integration with Black-box Attacks}
\label{app:integration}
In this section, we provide the detailed algorithm of our improved HR black-box attacks.

\subsection{High-resolution HSJ Attack}
\label{app:integration:hsj}


\begin{algorithm}[htb]
\caption{High-Resolution HSJ Attack (Simplified)}
\label{alg:hsj}
\begin{algorithmic}
\Require Scaling function $\Scale$, classifier $\ModelHR$, an image $X\in\SpaceHR$, iterations $T$, other parameters for HSJ attack.
\Ensure Perturbed image $X_t\in\SpaceHR$.
\State Initialize $\tilde{X}_0$ such that $\ModelHR(\tilde{X}_0)\neq \ModelHR(X)$.
\For{$t$ in $1, 2, ..., T$}
  \State $\triangleright$ {Binary Search}
  \State Find $X_t$ near the boundary between $X$ and $\tilde{X}_{t-1}$.
  \State $\triangleright$ \textbf{Scaling-aware Noise Sampling (\Cref{sec:method:sns})}
  \State Sample unit vectors $\s{U_1, U_2, ...}$ using \Cref{alg:sns}.
  \State $\triangleright$ {Gradient-direction Estimation}
  \State Estimate gradient direction $\bm{g}$ with $\s{U_1, U_2, ...}$.
  \State $\triangleright$ {Update Perturbed Image}
  \State Search the step size $\xi$.
  \State Set $\tilde{X}_t\gets X_t+\xi\cdot\bm{g}$.
\EndFor
\State Find $X_t$ near the boundary between $X$ and $\tilde{X}_{t-1}$.
\State Output $X_t$.
\end{algorithmic}
\end{algorithm}

\subsection{High-resolution Sign-OPT Attack}
\label{app:integration:opt}


\begin{algorithm}[htb]
\caption{High-Resolution SignOPT Attack (Simplified)}
\label{alg:opt}
\begin{algorithmic}
\Require Scaling function $\Scale$, classifier $\ModelHR$, an image $X\in\SpaceHR$, iterations $T$, other parameters for Sign-OPT attack.
\Ensure Adversarial direction $\bm{\theta}_t$.
\State Initialize adversarial direction $\bm{\theta}_0$.
\For{$t$ in $1, 2, ..., T$}
  \State $\triangleright$ \textbf{Scaling-aware Noise Sampling (\Cref{sec:method:sns})}
  \State Sample unit vectors $\s{U_1, U_2, ...}$ using \Cref{alg:sns}.
  \State $\triangleright$ {Gradient-direction Estimation}
  \State Estimate a better gradient $\hat{\bm{g}}$ with $\s{U_1, U_2, ...}$.
  \State $\triangleright$ {Update adversarial direction}
  \State Set $\bm{\theta}_{t}\gets\bm{\theta}_{t-1} - \eta\cdot\hat{\bm{g}}$.
  \State Search a point near the boundary along $\bm{\theta}_{t}$.
\EndFor
\State Output $\bm{\theta}_{t}$.
\end{algorithmic}
\end{algorithm}

\section{Circumventing Median Filtering Defense}
\label{app:median}
In this section, we provide more details of our circumvention of the median filtering defense, including its precise formulation, technical implementation, and additional empirical evaluations.

\subsection{Improve Gradient Estimation for Median}
\label{app:median:efficient}
For any input sequence $\bm{z}\in[0,1]^n$, the improved median function can be written as
\begin{equation} \label{eq:smooth-median}
\operatorname{improved-median}(\bm{z})\coloneqq\frac{\sum_{i=1}^nz_i\cdot \omega_i}{\sum_{i=1}^n \omega_i},
\end{equation}
where $\bm{\omega}\in\mathbb{R}^n$ is the weighting vector.

A useful weighting vector should satisfy two important properties: (1) it proportionally extends the gradient to non-median values; (2) it limits the number of changed values to mitigate the perturbation. We satisfy these two properties through quantile bounding and the absolute deviation to median, which define the weight as
\begin{equation}
\label{eq:attack:blackbox:weight}
\omega_i\coloneqq (1-\abs{z_i-\mathrm{median}(\bm{z})})\cdot\mathbbm{1}\s{z_{(a)}\leq z_i\leq z_{(b)}},
\end{equation}
where $z_{(a)}, z_{(b)}$ are the $a$-th and $b$-th quantile of scalar values in $\bm{z}$. We set $(a, b)$ to $(0.2, 0.8)$ based on an empirical evaluation in \Cref{app:median:quantile}. Intuitively, values that deviate more from the median are assigned smaller gradients, and the total number of changed values is limited if all values are close to the median.

We note that, however, the above approximation of median function may not be optimal; we leave better-optimized approximations as the future work.

\subsection{Approximation of Median's Gradient}
\label{app:median:quantile}

In \Cref{app:median:efficient}, we approximate the median function by ``trimmed and weighted average'' to provide a useful gradient for black-box attacks. To this end, we introduce the weight with quantile bounding, as defined in \Cref{eq:attack:blackbox:weight}. Here, we provide empirical evaluations of different choices of the quantile position $a$ and $b$ using our HR HSJ attack.

\begin{figure}[tb]
    \centering
    \includegraphics[width=0.75\linewidth]{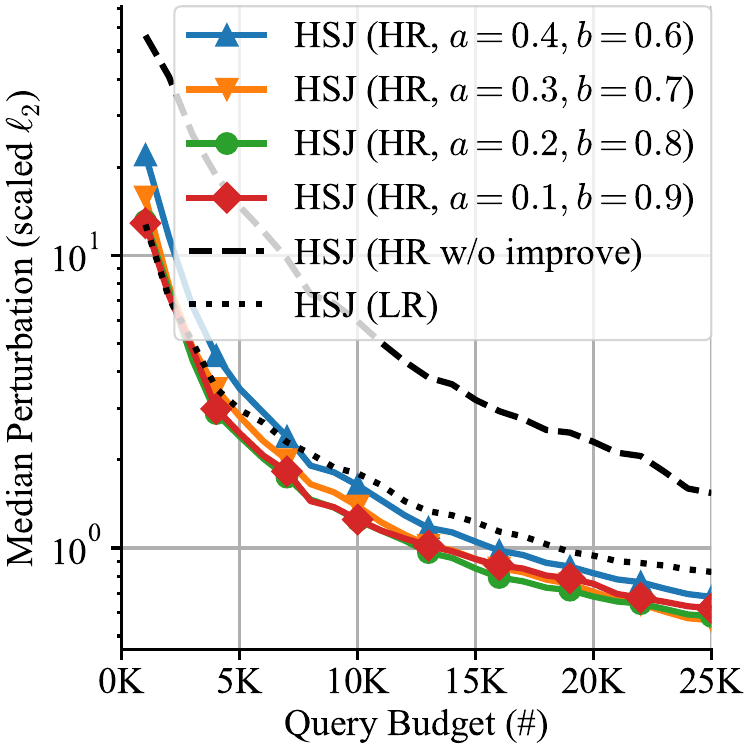}
    \caption{The performance of our HR HSJ (with improved median) under the median filtering defense and the LR HSJ. We show different quantile bounds in \Cref{eq:attack:blackbox:weight}.}
    \label{fig:sup:median}
\end{figure}

\Cref{fig:sup:median} shows that constrained bounds can result in suboptimal performance, such as $(0.4, 0.6)$ and $(0.3, 0.7)$. In contrast, relaxed bounds could obtain better performance, such as $(0.2, 0.8)$ and $(0.1, 0.9)$. We finally choose $(0.2, 0.8)$ as it obtains better performance when given lower query budgets (5K) and higher budgets (25K).

\subsection{Formalizing the Median Filtering Defense}
\label{app:median:formulation}

Our circumvention strategy in \Cref{sec:method:median} describes the high-level idea of improving the gradient estimation of the median \emph{function}. However, formulating the median filtering \emph{defense} is not straightforward because it is originally proposed as a \emph{selective} operation that only modifies vulnerable pixels, to which the scaling function gives high weights.

We will first show how to formulate the selective filtering defense as a \emph{masked pooling layer} with a boolean mask that represents vulnerable pixels. After that, we show a simple strategy to identify these vulnerable pixels.

\paragraph{Masked Pooling Layer.}
We describe the pooling layer as a convolution, as the pooling layer works like a discrete convolution but replaces the linear kernel with some other function~\cite{pooling}. This allows us to represent the defense as
\begin{equation}\label{eq:attack:pooling}
    \Defense(X) \coloneqq p(X) \cdot \mathtt{mask} + X \cdot (1 - \mathtt{mask}),
\end{equation}
where $p$ is the pooling function, $\mathtt{mask}$ is a boolean mask with $1$ denoting vulnerable pixels. The pooling function is given as
$p(X) \coloneqq X \star \Kernel$, 
where $\star$ denotes 2D convolution with reflect padding to keep the same shape, and $\Kernel$ denotes the filter function determined by the defense. In the case of the median filtering defense, we set $\Kernel$ to the  improved median function in \Cref{eq:smooth-median} in the backward pass, and switch to the standard median function in the forward pass.

\paragraph{Identifying Vulnerable Pixels.}
We then explain how to determine the boolean mask in \Cref{eq:attack:pooling}.
For any fixed scaling algorithm, we can write the scaling function as a matrix multiplication like \Cref{eq:scaling:matrix}. By setting all entries in $\bm{x}\in\SpaceLR$ to one and solving for $X\in\SpaceHR$, we have
\begin{equation}\label{eq:attack:scaling.pinv}
    X^* = L^+ \times D \times R^+ \in\SpaceHR,
\end{equation}
where $L^+$ and $R^+$ are the pseudo-inverse \cite{pseudo-inverse} of $L$ and $R$.
Conceptually, this recovers the element-wise weight of each pixel in the source image during scaling.
That is, every non-zero entry in $X^*$ indicates a vulnerable pixel in the source image; we thus determine the boolean mask as
\begin{equation}\label{eq:attack:scaling.mask}
    \mathtt{mask} = \mathbbm{1}\s{X^*\neq0},
\end{equation}
where the indicator function $\mathbbm{1}$ and operator $\neq$ are all computed in element-wise. 

In summary, we have formalized the median filtering defense as a masked pooling layer in \Cref{eq:attack:pooling}. This formulation allows us to apply SNS in \Cref{sec:method:sns} with a precise definition of $(\Defense\circ\Scale)$.

\section{Experimental Details}
\label{app:setup}
In this section, we provide more details of our evaluation setup. We run all experiments on 8 Nvidia RTX 2080 Ti GPUs, each with 11 GB memory.

\subsection{Datasets and Models}
\label{app:setup:models}

We use two datasets in our evaluation: ImageNet~\cite{imagenet} and CelebA~\cite{celeba}.

\textbf{ImageNet}. We randomly choose 1,000 images larger than $672\times 672$ and downscale them to $672\times 672$. The target model is a ResNet-50~\cite{resnet} model pre-trained by TorchVision\footnote{\url{https://pytorch.org/vision/stable/models.html}}, which attains 76.13\% Top-1 accuracy and 92.86\% Top-5 accuracy on ImageNet. We discard source images that are mis-classified before running the attack. 

In the white-box setting, we use a ResNet-50 model adversarially trained by \citet{robustness}, which attains 57.90\% Top-1 accuracy on benign inputs and 35.09\% Top-1 accuracy under PGD attack with 100 steps and an \LL{2}-norm budget of 3. We only choose correctly classified images to avoid the artifacts of its slightly lower benign accuracy.

\textbf{CelebA}. We randomly choose 1,000 images and rescale their faces to $672\times 672$. For simplicity, we pre-cropped the facial images and directly evaluate LR and HR attacks on such images, but it is straightforward to adopt the complete pre-processing pipeline that includes facial extraction, as this procedure is not randomized. The target model is a pre-trained ResNet-34 model that accepts facial images of size $224\times 224$ and predicts the \texttt{Mouth\_Slightly\_Open} attribute with 92.4\% accuracy.

\subsection{Image Analysis API}
\label{app:setup:api}

The Tencent Image Analysis API accepts a variety of images and returns Top-5 labels (with probability scores) that best describe the image. This API uses OpenCV's linear scaling as inferred by \citet{scaling2019}. 

We define the ground-truth label as the benign input's Top-1 label; we only consider benign inputs whose Top-1 score is above 50\%. A successful attack should decrease the true label's score to below 10\%. Our attacks did not leverage these scores; they only have access to the final decision.
When running transfer-based attacks on this API, we adopt the robust model in \Cref{app:setup:models} as the surrogate model; attacks on a non-robust model cannot transfer to this API.

\subsection{Attacks}
\label{app:setup:attacks}

We use HSJ~\cite{hopskip}, C\&W\cite{carlini2019evaluating}, and PGD~\cite{pgd} attacks implemented by Adversarial Robustness Toolbox\footnote{\url{https://github.com/Trusted-AI/adversarial-robustness-toolbox}}~\cite{art2018}. For the Sign-OPT attack~\cite{signoptattack}, we use its official implementation\footnote{\url{https://github.com/cmhcbb/attackbox}}. Particularly, we did not change the default parameters used in black-box attacks; all optimization parameters are fixed to the official recommendation.

For the C\&W attack, we set the binary search step to 20 with a maximum of 1,000 iterations. The confidence parameter $\kappa$ is set to $\s{0, 1, ..., 10}$.
For the PGD attack, we set the number of steps to 100 with \LL{2}-norm budget $\epsilon=\s{1,2,\dots20}$ and step size $0.1\times\epsilon$.

\section{Exploiting Vulnerabilities Sequentially}
\label{app:sequential}

In \Cref{sec:eval:none}, we have shown how to \emph{jointly} exploit vulnerabilities in upstream scaling and downstream classifier. However, we note that it is also possible to \emph{sequentially} exploit these two vulnerabilities. For example, the attacker can deploy the conventional black-box attacks on the downstream model, and leverage the image-scaling attack to hide the adversarial example within its original clean image. This requires solving \Cref{eq:scaling-attack}, where $T$ is the standard adversarial example generated by a black-box attack on the downstream model.

However, we note that the sequential attack is suboptimal. This attack only finds an HR image whose downscaled version is close enough to the given LR adversarial example; it cannot guarantee that the obtained HR image is still adversarial after downscaling. As black-box adversarial examples are typically near the decision boundary, the final solution may still lie in the correct label's decision area even if it is close enough to the given adversarial example. This problem is more severe when it is hard to precisely invert the median filtering defense. 

\Cref{fig:exp:gen-vs-hide} compares the performance of our joint attack and the alternative sequential attack (all based on HSJ). The sequential attack injects adversarial examples from LR HSJ to their HR source images. \Cref{fig:exp:gen-vs-hide:1} shows that jointly attacking the pipeline uses the query budget more efficiently; the 5K joint attack even beats the 20K sequential attack. \Cref{fig:exp:gen-vs-hide:2} shows that the sequential attack is suboptimal under the median defense; it becomes worse when the target adversarial example was obtained with more queries (thus, more sensitive to the imprecise results from image-scaling attacks). We thus focus on the more practical joint attack, which directly optimizes to attack the entire ML system.

Finally, we conjecture that the sequential attack only works when the given target image can induce a misclassification with high probability, such as an image from another class (like the standard image-scaling attack) or an adversarial example generated by the C\&W attack with high confidence.

\section{Analyzing Randomized Filtering Defense}
\label{app:random}

In the following arguments, we show that the randomized filtering defense, when viewed jointly with the scaling function, can be regarded as a uniform scaling procedure.

Without loss of generality, we study randomized filtering over a $3\times 3$ window $\bm{w}$ in the source image $S$ and an arbitrary convolution kernel $\bm{k}$. We also pad the input properly so the window $\bm{w}$ is always surrounded by other pixels. Since both the scaling and filtering functions can be written as a convolution, we restate the defended scaling $D=(\Scale\circ\Defense)(S)$ over a window $\bm{w}$ with output pixel $d_{2,2}$ as
\begin{equation*}\label{eq:pooling:random:proof}
\begin{aligned}
d_{2,2} &= \bm{w} \star \Kernel_\mathrm{rnd} \star \bm{k} \\
&=
\begin{bmatrix}
w_{1,1} & w_{1,2} & w_{1,3} \\
w_{2,1} & w_{2,2} & w_{2,3} \\
w_{3,1} & w_{3,2} & w_{3,3}
\end{bmatrix}
\star
f_\mathrm{rnd}
\star
\begin{bmatrix}
k_{1,1} & k_{1,2} & k_{1,3} \\
k_{2,1} & k_{2,2} & k_{2,3} \\
k_{3,1} & k_{3,2} & k_{3,3}
\end{bmatrix},
\end{aligned}
\end{equation*}
where the randomized filter $\Kernel_\mathrm{rnd}$ randomly picks a pixel from the $3\times3$ window $w$ with probability $1/9$.

We study the central pixel $w_{2,2}$ and its weight $w^\prime_{2,2}$ during this defended scaling.
First, the randomized filtering slides a $3\times3$ window around each pixel $w_{i,j}$ and randomly changes its value to $w_{2,2}$ with a probability $\mathrm{Pr}[w_{i,j}\gets w_{2,2}]=1/9$.
Second, the scaling algorithm gives the weight $k_{i,j}$ to each pixel $w_{i,j}$. Since the pixel $w_{i,j}$ could hold the value of $w_{2,2}$, the overall weight of $w_{2,2}$ can be described as $\mathrm{Pr}[w^\prime_{2,2}\gets k_{i,j}]=1/9$.
Thus, we can write the expected value of the weight $w^\prime_{2,2}$ as
\begin{equation}\label{eq:pooling:random:kernel:single}
\EE_{\Kernel\sim\mathcal{K}}[w^\prime_{2,2}]
=\sum_{1\leq i,j\leq 3} \frac{1}{9} \cdot k_{i,j}
= \frac{1}{9},
\end{equation}
where $\mathcal{K}$ is the filter space determined by $\Kernel_\mathrm{rnd}$ and we have assumed a normalized scaling kernel $\bm{k}$. This shows that the pixel $w_{2,2}$ is given a uniform weight in expectation. Extending to other pixels, we have:
\begin{equation}
\EE_{\Defense\sim\mathcal{H}}[(\Scale\circ\Defense)(S)] = S\star \bm{k}_\mathrm{u},
\end{equation}
where $\mathcal{H}$ is the space of defense functions chosen by the randomized filtering defense and $\bm{k}_\mathrm{u}$ is the uniform area scaling kernel defined in \Cref{sec:background:scaling-defenses:robust}.

\section{High-Resolution White-box Attacks}
\label{app:whitebox}

In this section, we provide details of extending white-box attacks~\cite{carlini2019evaluating,pgd} to the full ML pipeline. This is useful for transfer-based black-box attacks as well as evaluating the worst-case robustness of ML pipeline and image-scaling defenses. Since white-box attacks can easily circumvent defenses using BPDA~\cite{obfuscated} and EOT~\cite{eot}, this evaluation is only for the completeness of this paper and does not claim any technical novelty.

\subsection{Formulating White-box Attacks}
\label{app:whitebox:formulation}

Formulating white-box attacks is straightforward by viewing the entire ML pipeline as a sequential model. For example, the objective function of C\&W attack becomes
\begin{equation}\label{eq:whitebox:cw}
\begin{aligned}
\min &\ \norm{\Delta}_2 + c\cdot (\ModelLR^\prime\circ \Scale\circ\Defense)(X+\Delta) \\
\mathrm{s.t.} &\ X+\Delta \in \SpaceHR,
\end{aligned}
\end{equation}
where $\ModelLR^\prime$ is the loss function quantifying the confidence of the model $\ModelLR$'s ground-truth prediction, $X$ is the HR source image, and $\Delta$ is the HR adversarial perturbation. Similarly, the objective function of PGD attack becomes
\begin{equation}\label{eq:whitebox:pgd}
\begin{aligned}
\max &\ J\p[\big]{(\ModelLR\circ\Scale\circ\Defense)(X+\Delta), y^*} \\
\mathrm{s.t.} &\ \norm{\Delta}_2\leq\epsilon,
\end{aligned}
\end{equation}
where $J(\cdot)$ is the cross entropy loss, $y^*$ is the ground truth label of $X$, and $\epsilon$ is the specified perturbation budget.

\begin{figure}[t]
\centering

\begin{subfigure}[t]{0.32\linewidth}
	\includegraphics[width=\linewidth]{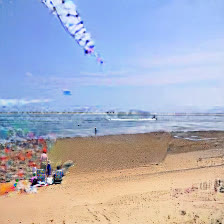}
	\caption{\centering C\&W Vanilla  (scaled $\ell_2 = 17.44$)}
\end{subfigure}
\hfill
\begin{subfigure}[t]{0.32\linewidth}
	\includegraphics[width=\linewidth]{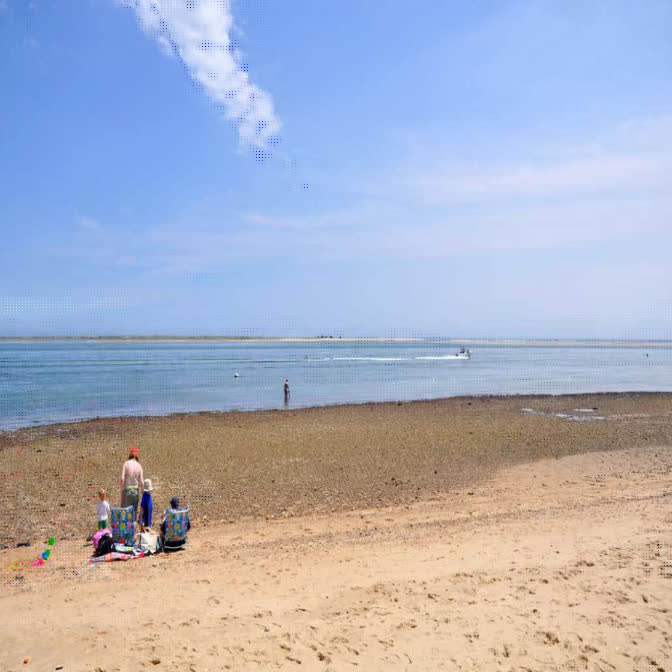}
	\caption{\centering C\&W Scaling   (scaled $\ell_2 = 5.84$)}
\end{subfigure}
\hfill
\begin{subfigure}[t]{0.32\linewidth}
	\includegraphics[width=\linewidth]{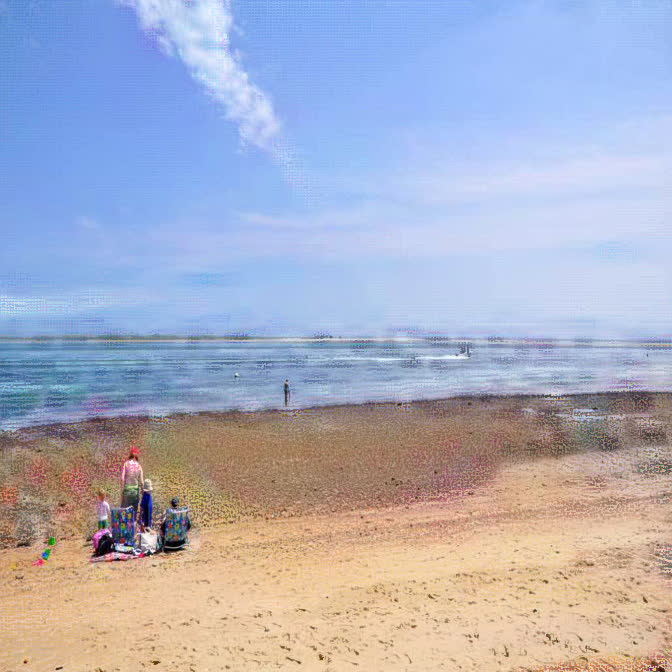}
	\caption{\centering C\&W Medain (scaled $\ell_2 = 14.4$)}
\end{subfigure}

\caption{Adversarial examples from (a) C\&W, (b) HR C\&W, and (c) HR C\&W under the median filtering defense. The confidence is set to $\kappa=2$. The shape is $224\times 224$ for (a), and $672\times 672$ for (b) and (c). HR C\&W attack produces less perturbation even under the median defense.}
\label{fig:attack:cw}
\end{figure}

\subsection{Evaluating Image-scaling Defenses}
\label{app:whitebox:evaluate}

\subsubsection{Pre-processing Defenses}
From the perspective of a whole ML pipeline, pre-processing defenses for the scaling function can be circumvented by BPDA~\cite{obfuscated} in white-box attacks. However, as we find the gradient of our masked pooling layer formulation in \Cref{app:median:formulation} to be useful, we will keep the median filtering formulated as \Cref{eq:attack:pooling} for consistency. Adversarial examples produced by HR C\&W attacks are shown in \Cref{fig:attack:cw}.

We use HR C\&W and PGD attacks in \Cref{app:whitebox:formulation} to evaluate the robustness under the confidence and perturbation constraint, respectively. Similar to the setting in \Cref{sec:eval:preprocessing}, we generate two sets of adversarial examples using the LR and HR attacks, respectively.

\begin{figure}[t]
\centering
    \begin{subfigure}[t]{0.49\linewidth}
        \includegraphics[width=\linewidth]{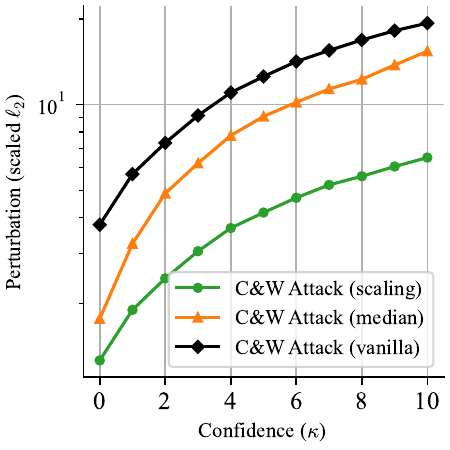}
        \caption{Perturbation (scaled $\ell_2$)}
        \label{fig:exp:cw}
    \end{subfigure}
    \begin{subfigure}[t]{0.49\linewidth}
        \includegraphics[width=\linewidth]{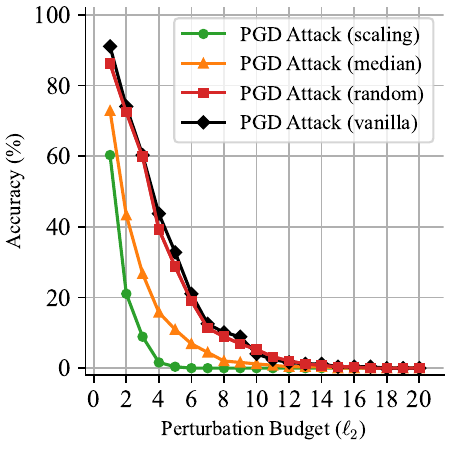}
        \caption{Accuracy (\%)}
        \label{fig:exp:pgd}
    \end{subfigure}
    \caption{Comparison of HR white-box attacks under different constraints and prevention scaling defenses. Only the randomized filtering defense is robust.}
    \label{fig:exp:whitebox}
\end{figure}

\Cref{fig:exp:whitebox} shows the performance of HR white-box attacks when attacking the entire ML system pipeline. Overall, HR attacks are able to gain incentives under no scaling defense or the median filtering defense. In \Cref{fig:exp:cw}, the HR C\&W attack was able to achieve the same confidence with lower perturbation. In \Cref{fig:exp:pgd}, the HR PGD attack was able to decrease more accuracy with the same perturbation budget.

The only exception is the randomized filtering defense in \Cref{fig:exp:pgd}, which successfully protects the ML system's robustness against threats from the scaling procedure. We show that scaling algorithms adopting uniform kernels or dynamic kernel widths~\cite{scaling2020} are robust as well in \Cref{app:whitebox:robust}.

\subsubsection{Detection Defenses}
Although we empirically observe that HR white-box attacks evade all existing detection defenses out of the box, we note that a learning-based detector may still work. In that case, one could adaptively evaluate the detector's effectiveness using the following approach. Given any detection function $d$, we first choose a loss function $L$ so that $L(X+\Delta)$ is minimized when the detection $d(X+\Delta)$ is incorrect. We then add the loss function $L$ as a regularizer to our objective functions. For instance, \Cref{eq:whitebox:pgd} would become:
\begin{equation*}\label{eq:attack:generate:new}
\begin{aligned}
\max &\ J\p[\big]{(\ModelLR\circ\Scale\circ\Defense)(X+\Delta), y} - \gamma\cdot L(X+\Delta) \\
\mathrm{s.t.} &\ \norm{\Delta}_2\leq\epsilon,
\end{aligned}
\end{equation*}
where $\gamma$ is the hyper-parameter that controls the weight of the added regularizer. This approach is similar to the attack from \citet{frequency-detection-attack} against a learning-based deep-fake detector~\cite{frequency-detection}, which also detects the artifacts of deep-fakes in the spectrum domain. We leave the investigation of such attacks to future work.

\begin{figure}[tb]
    \centering
    \begin{subfigure}[t]{0.48\linewidth}
        \includegraphics[width=\linewidth]{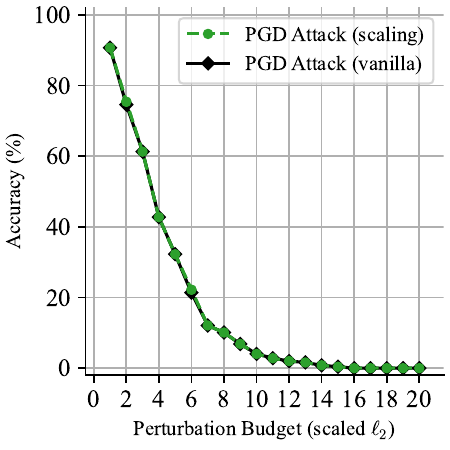}
        \caption{CV Area}
    \end{subfigure}
    \begin{subfigure}[t]{0.48\linewidth}
        \includegraphics[width=\linewidth]{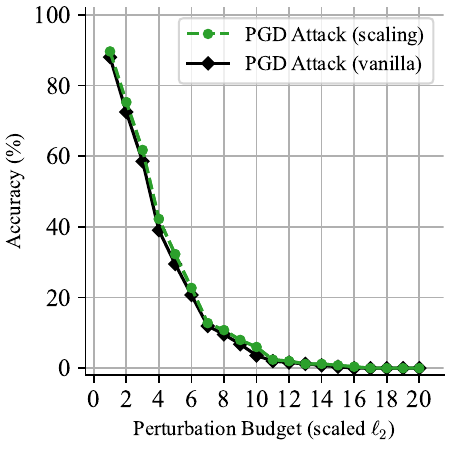}
        \caption{PIL Linear}
    \end{subfigure}
    \caption{Compare the performance of vanilla and HR PGD attacks on scaling algorithms that are robust by design. This verifies the robustness of such scaling algorithms.}
    \label{fig:exp:pgd:robust}
\end{figure}

\subsubsection{Robust Scaling Algorithms}
\label{app:whitebox:robust}
Recall that \citet{scaling2020} have identified a few scaling algorithms that are robust against the scaling attack (see \Cref{sec:background:scaling-defenses:robust}). We also evaluate their robustness against the HR image-scaling attack. In \Cref{fig:exp:pgd:robust}, we report the evaluation of scaling algorithms that adopt uniform kernels (CV Area) or dynamic kernel widths (PIL Linear). As evident from the plots, HR PGD attacks cannot exploit these scaling algorithms to improve the vanilla one.  This verifies the robustness of known-robust scaling algorithms in a setting where the attacker jointly targets the whole ML system pipeline.

\section{Black-box Adversarial Examples by HSJ}
\label{app:blackboxexample}
\begin{figure*}
\centering
\captionsetup[subfigure]{labelformat=empty}

\begin{subfigure}[t]{0.12\linewidth}
    \includegraphics[width=\linewidth]{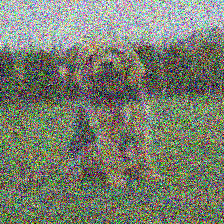}
    \caption{$\ell_2 = 102.97$}
\end{subfigure}
\hspace{0.1em}
\begin{subfigure}[t]{0.12\linewidth}
    \includegraphics[width=\linewidth]{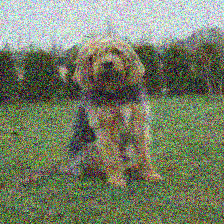}
    \caption{$\ell_2 = 65.52$}
\end{subfigure}
\hspace{0.1em}
\begin{subfigure}[t]{0.12\linewidth}
    \includegraphics[width=\linewidth]{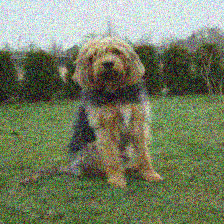}
    \caption{$\ell_2 = 44.56$}
\end{subfigure}
\hspace{0.1em}
\begin{subfigure}[t]{0.12\linewidth}
    \includegraphics[width=\linewidth]{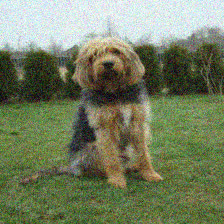}
    \caption{$\ell_2 = 29.36$}
\end{subfigure}
\hspace{0.1em}
\begin{subfigure}[t]{0.12\linewidth}
    \includegraphics[width=\linewidth]{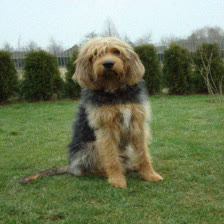}
    \caption{$\ell_2 = 6.53$}
\end{subfigure}
\hspace{0.1em}
\begin{subfigure}[t]{0.12\linewidth}
    \includegraphics[width=\linewidth]{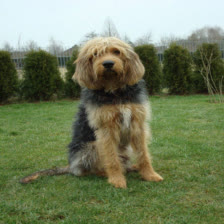}
    \caption{$\ell_2 = 3.13$}
\end{subfigure}

\vspace{0.5em}

\begin{subfigure}[t]{0.12\linewidth}
    \includegraphics[width=\linewidth]{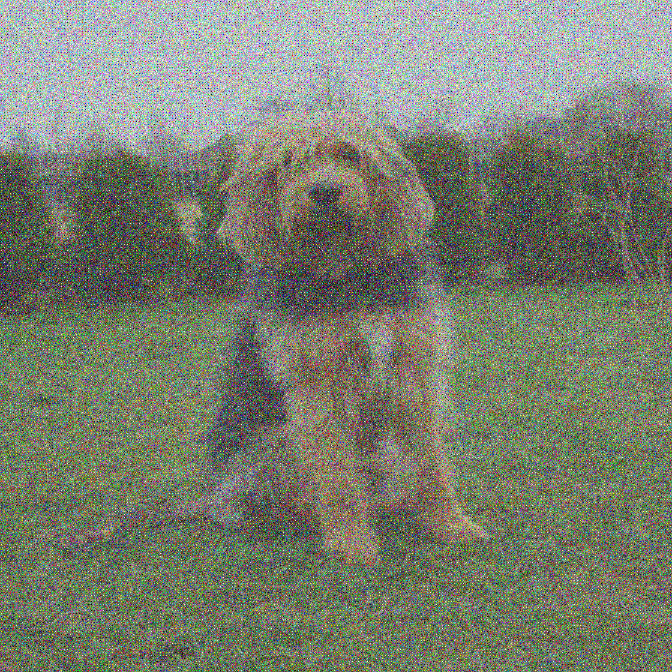}
    \caption{$\ell_2 = \mathbf{92.00}$}
\end{subfigure}
\hspace{0.1em}
\begin{subfigure}[t]{0.12\linewidth}
    \includegraphics[width=\linewidth]{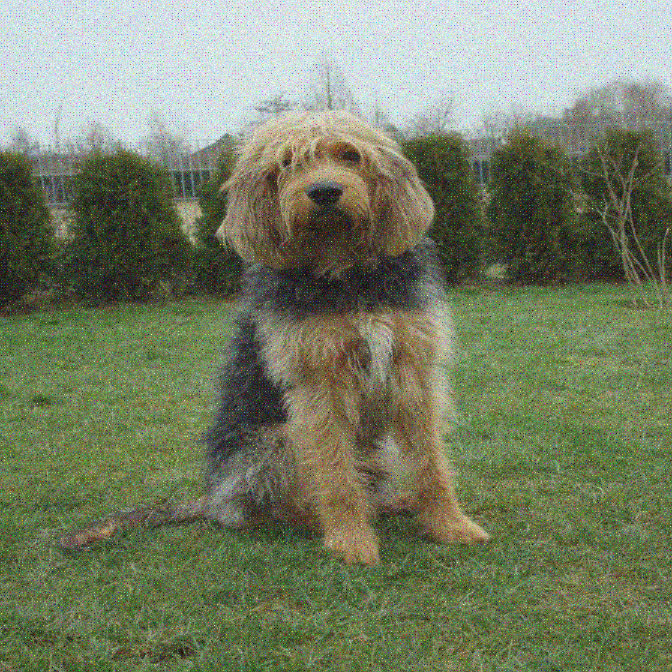}
    \caption{$\ell_2 = \mathbf{41.31}$}
\end{subfigure}
\hspace{0.1em}
\begin{subfigure}[t]{0.12\linewidth}
    \includegraphics[width=\linewidth]{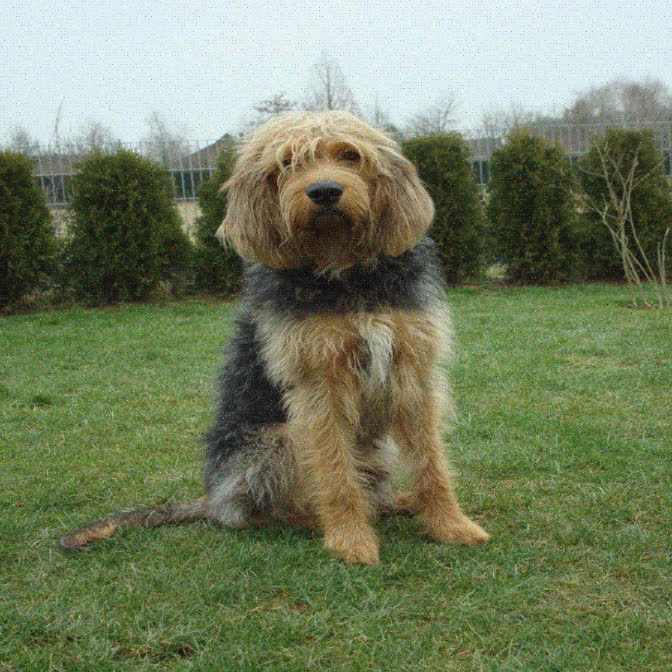}
    \caption{$\ell_2 = \mathbf{19.13}$}
\end{subfigure}
\hspace{0.1em}
\begin{subfigure}[t]{0.12\linewidth}
    \includegraphics[width=\linewidth]{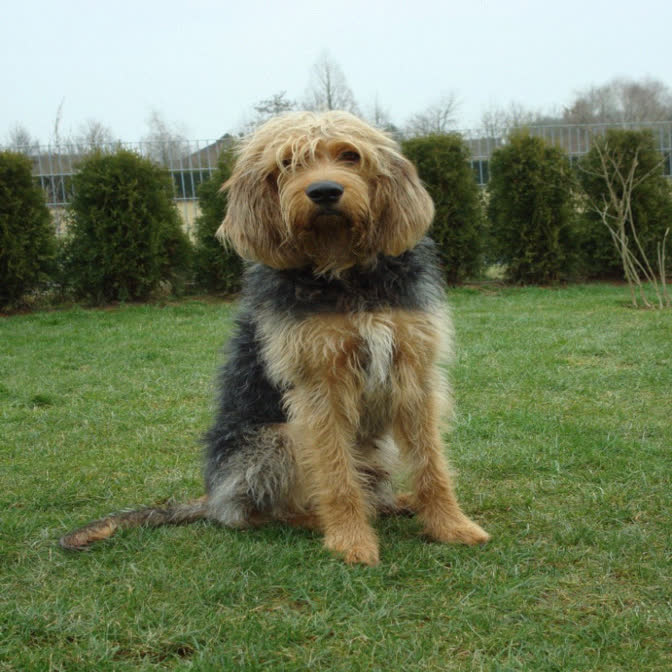}
    \caption{$\ell_2 = \mathbf{5.12}$}
\end{subfigure}
\hspace{0.1em}
\begin{subfigure}[t]{0.12\linewidth}
    \includegraphics[width=\linewidth]{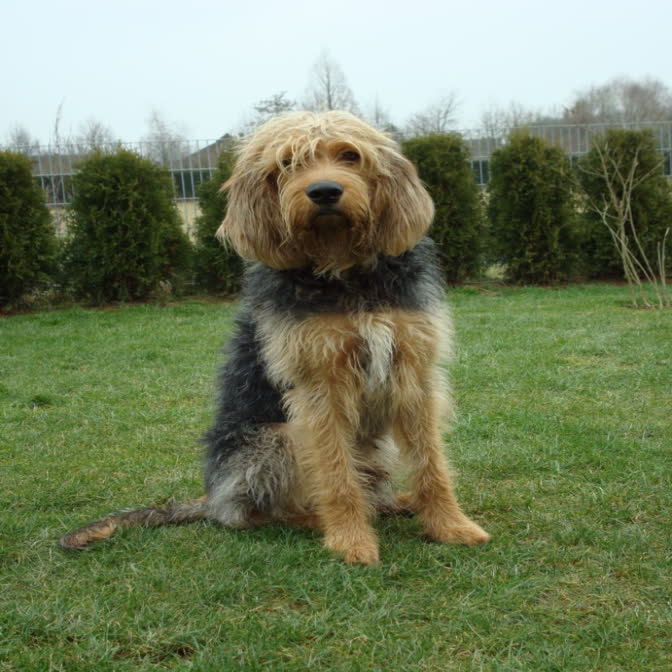}
    \caption{$\ell_2 = \mathbf{0.50}$}
\end{subfigure}
\hspace{0.1em}
\begin{subfigure}[t]{0.12\linewidth}
    \includegraphics[width=\linewidth]{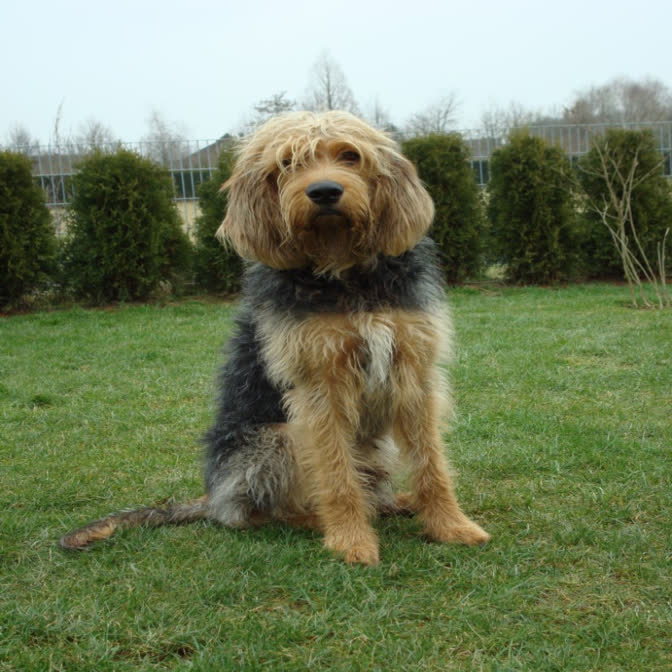}
    \caption{$\ell_2 = \mathbf{0.28}$}
\end{subfigure}

\vspace{0.5em}

\begin{subfigure}[t]{0.12\linewidth}
    \includegraphics[width=\linewidth]{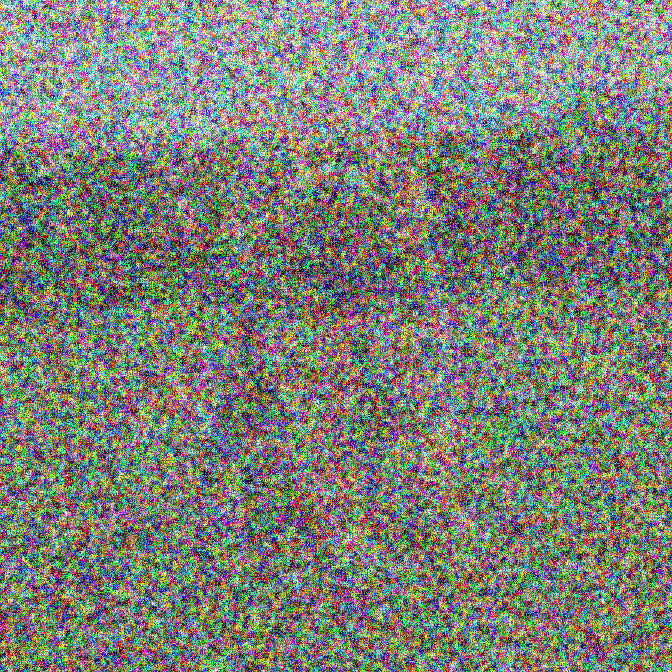}
    \caption{$\ell_2 = 141.19$}
\end{subfigure}
\hspace{0.1em}
\begin{subfigure}[t]{0.12\linewidth}
    \includegraphics[width=\linewidth]{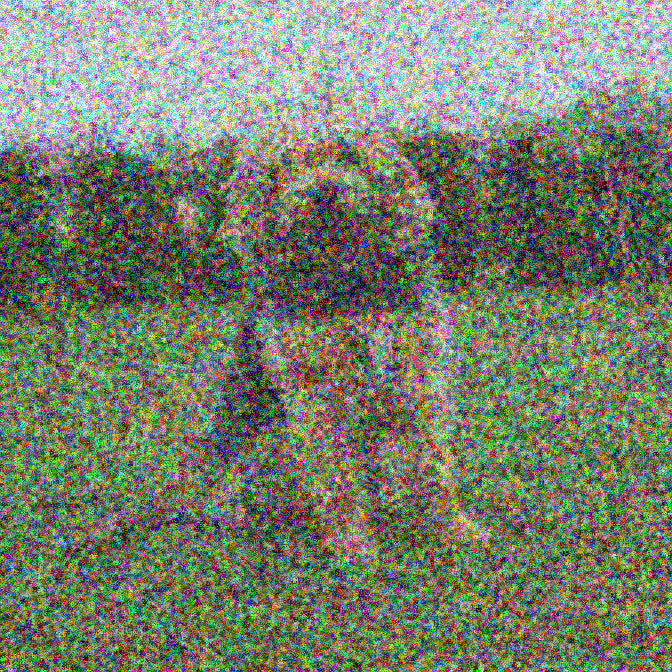}
    \caption{$\ell_2 = 94.72$}
\end{subfigure}
\hspace{0.1em}
\begin{subfigure}[t]{0.12\linewidth}
    \includegraphics[width=\linewidth]{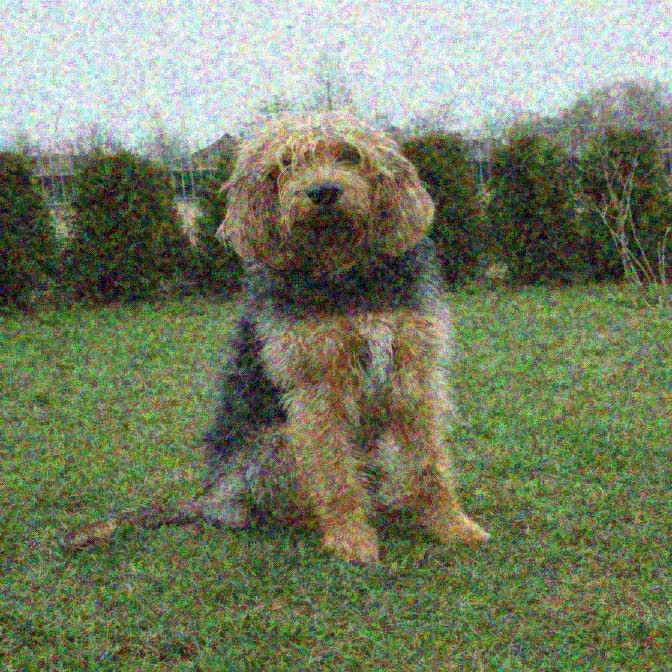}
    \caption{$\ell_2 = \mathbf{40.54}$}
\end{subfigure}
\hspace{0.1em}
\begin{subfigure}[t]{0.12\linewidth}
    \includegraphics[width=\linewidth]{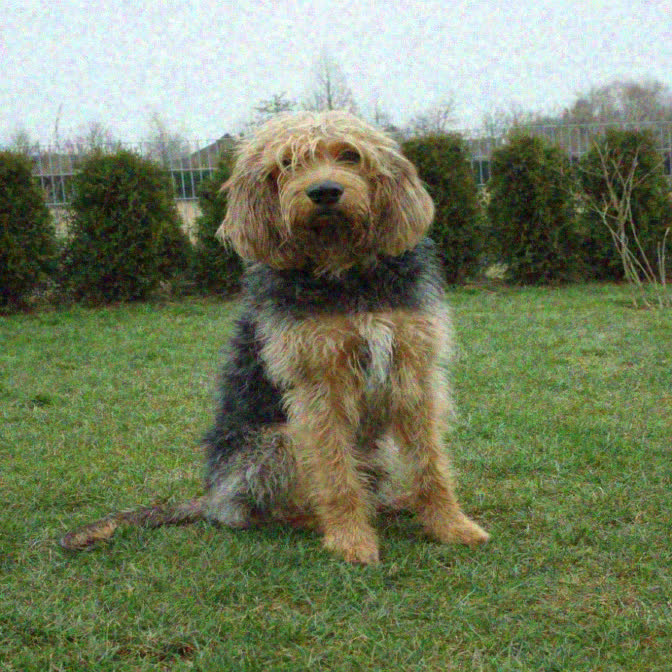}
    \caption{$\ell_2 = \mathbf{17.68}$}
\end{subfigure}
\hspace{0.1em}
\begin{subfigure}[t]{0.12\linewidth}
    \includegraphics[width=\linewidth]{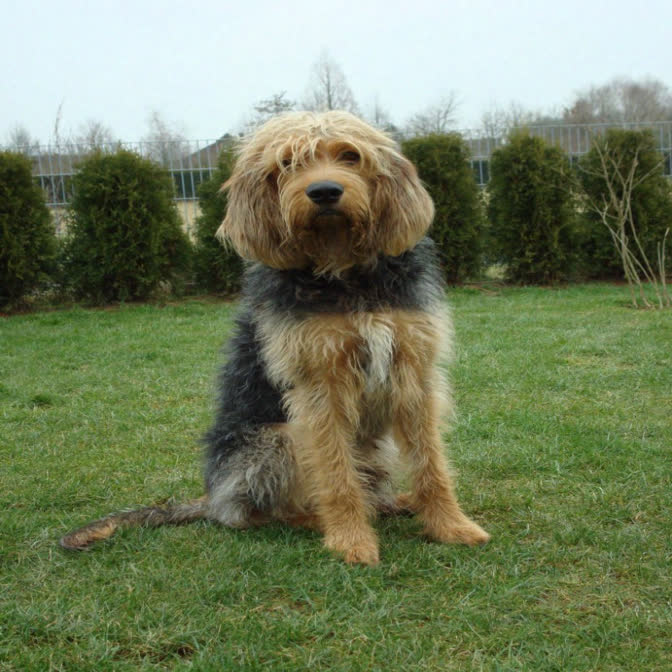}
    \caption{$\ell_2 = \mathbf{2.85}$}
\end{subfigure}
\hspace{0.1em}
\begin{subfigure}[t]{0.12\linewidth}
    \includegraphics[width=\linewidth]{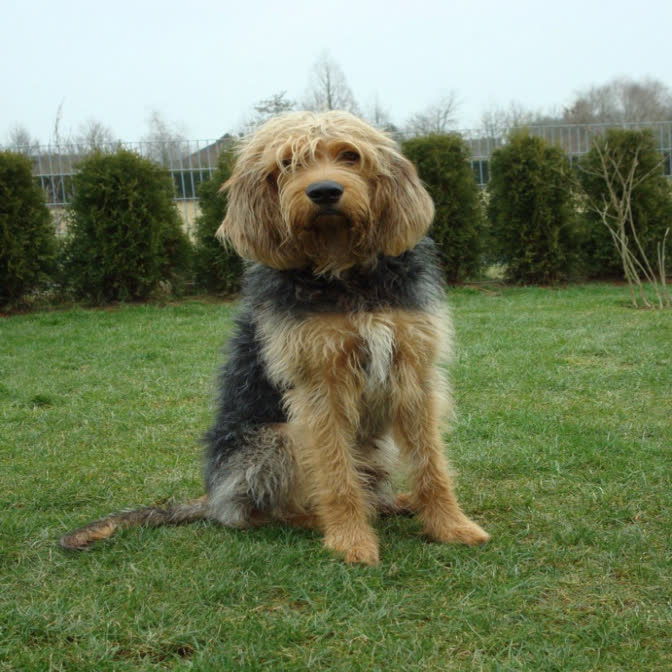}
    \caption{$\ell_2 = \mathbf{1.45}$}
\end{subfigure}

\vspace{0.5em}

\begin{subfigure}[t]{0.12\linewidth}
    \includegraphics[width=\linewidth]{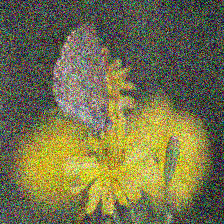}
    \caption{$\ell_2 = 77.26$}
\end{subfigure}
\hspace{0.1em}
\begin{subfigure}[t]{0.12\linewidth}
    \includegraphics[width=\linewidth]{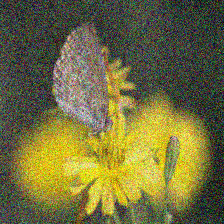}
    \caption{$\ell_2 = 52.14$}
\end{subfigure}
\hspace{0.1em}
\begin{subfigure}[t]{0.12\linewidth}
    \includegraphics[width=\linewidth]{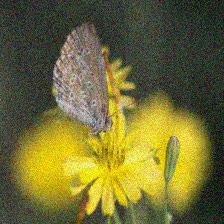}
    \caption{$\ell_2 = 29.18$}
\end{subfigure}
\hspace{0.1em}
\begin{subfigure}[t]{0.12\linewidth}
    \includegraphics[width=\linewidth]{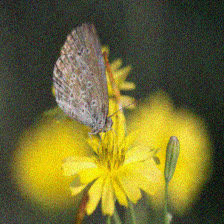}
    \caption{$\ell_2 = 18.77$}
\end{subfigure}
\hspace{0.1em}
\begin{subfigure}[t]{0.12\linewidth}
    \includegraphics[width=\linewidth]{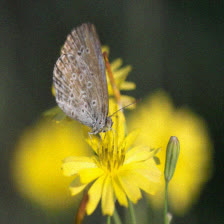}
    \caption{$\ell_2 = 5.90$}
\end{subfigure}
\hspace{0.1em}
\begin{subfigure}[t]{0.12\linewidth}
    \includegraphics[width=\linewidth]{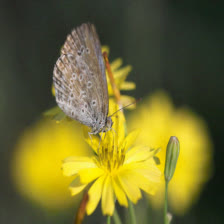}
    \caption{$\ell_2 = 3.23$}
\end{subfigure}

\vspace{0.5em}

\begin{subfigure}[t]{0.12\linewidth}
    \includegraphics[width=\linewidth]{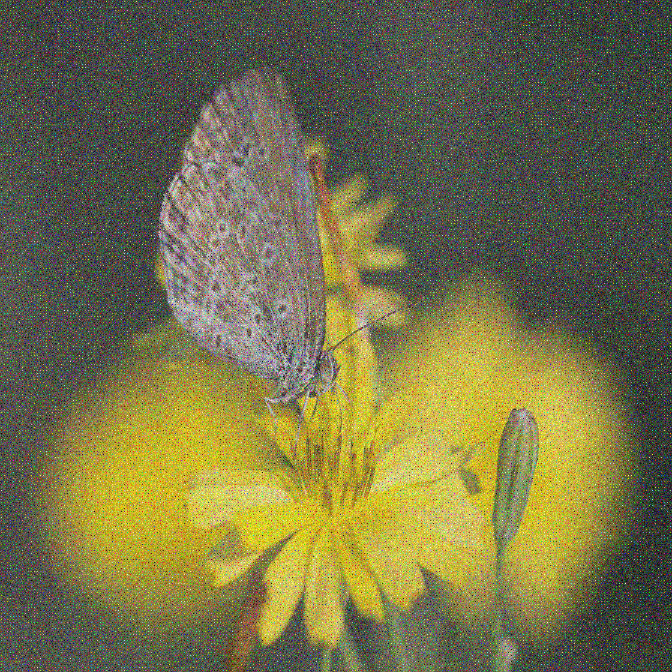}
    \caption{$\ell_2 = \mathbf{67.02}$}
\end{subfigure}
\hspace{0.1em}
\begin{subfigure}[t]{0.12\linewidth}
    \includegraphics[width=\linewidth]{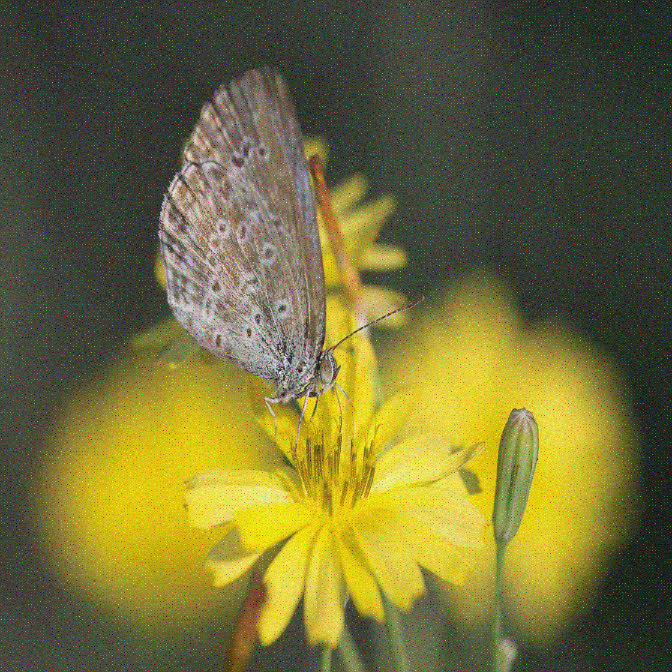}
    \caption{$\ell_2 = \mathbf{30.75}$}
\end{subfigure}
\hspace{0.1em}
\begin{subfigure}[t]{0.12\linewidth}
    \includegraphics[width=\linewidth]{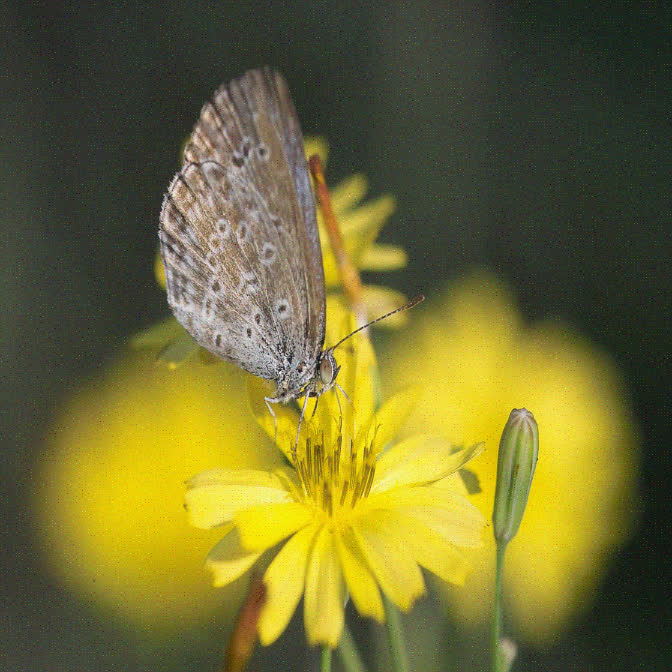}
    \caption{$\ell_2 = \mathbf{14.41}$}
\end{subfigure}
\hspace{0.1em}
\begin{subfigure}[t]{0.12\linewidth}
    \includegraphics[width=\linewidth]{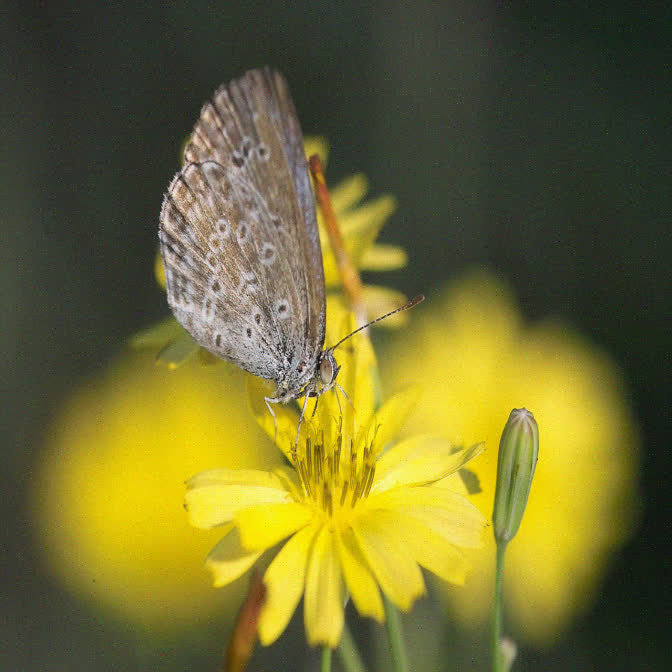}
    \caption{$\ell_2 = \mathbf{9.58}$}
\end{subfigure}
\hspace{0.1em}
\begin{subfigure}[t]{0.12\linewidth}
    \includegraphics[width=\linewidth]{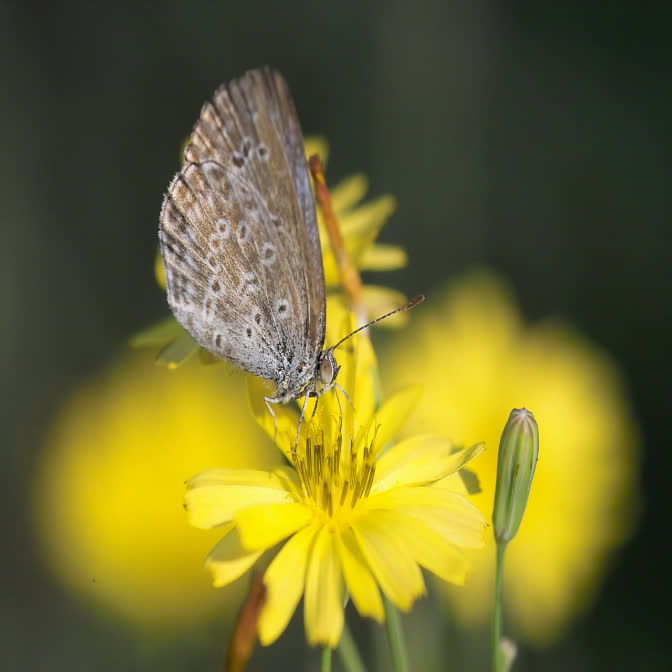}
    \caption{$\ell_2 = \mathbf{2.04}$}
\end{subfigure}
\hspace{0.1em}
\begin{subfigure}[t]{0.12\linewidth}
    \includegraphics[width=\linewidth]{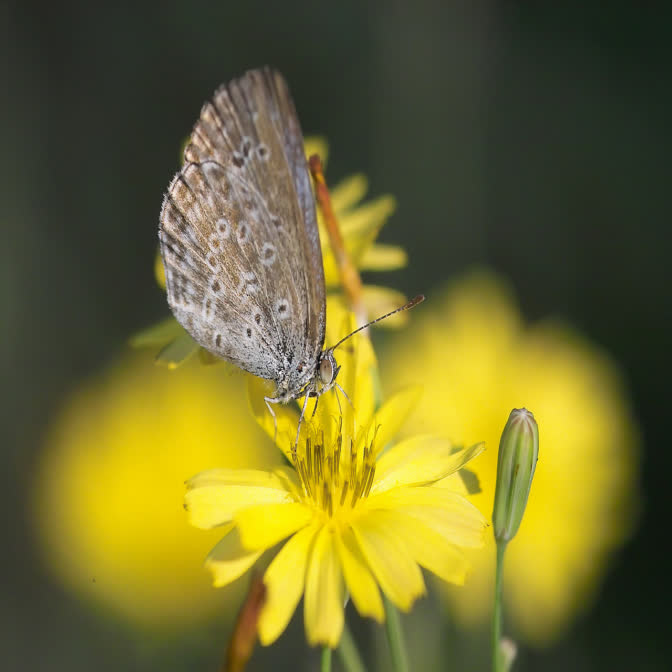}
    \caption{$\ell_2 = \mathbf{1.08}$}
\end{subfigure}

\vspace{0.5em}

\begin{subfigure}[t]{0.12\linewidth}
    \includegraphics[width=\linewidth]{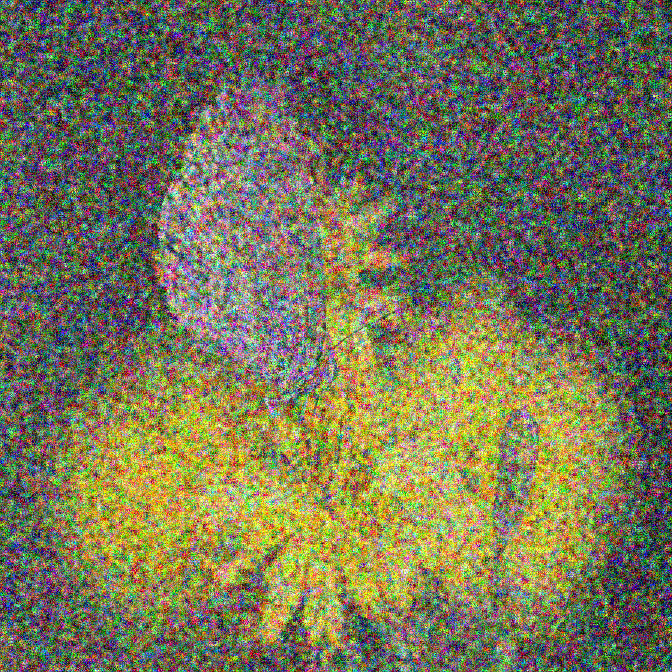}
    \caption{$\ell_2 = 104.79$}
\end{subfigure}
\hspace{0.1em}
\begin{subfigure}[t]{0.12\linewidth}
    \includegraphics[width=\linewidth]{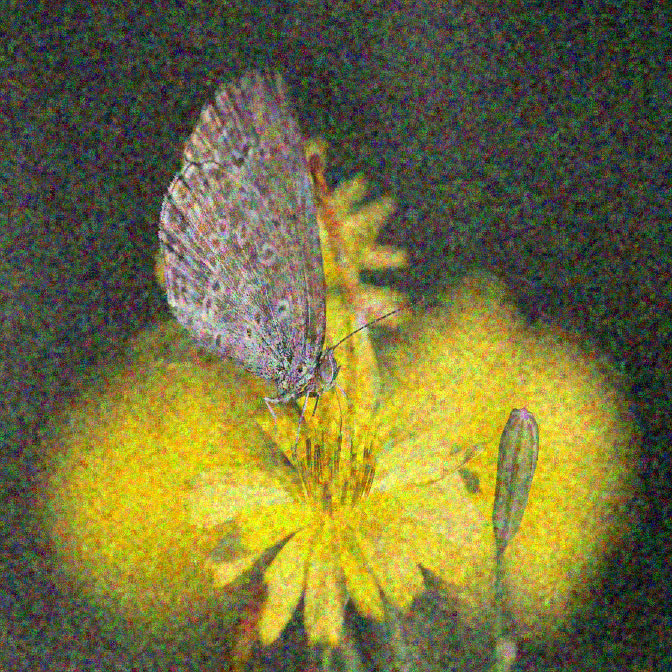}
    \caption{$\ell_2 = \mathbf{44.41}$}
\end{subfigure}
\hspace{0.1em}
\begin{subfigure}[t]{0.12\linewidth}
    \includegraphics[width=\linewidth]{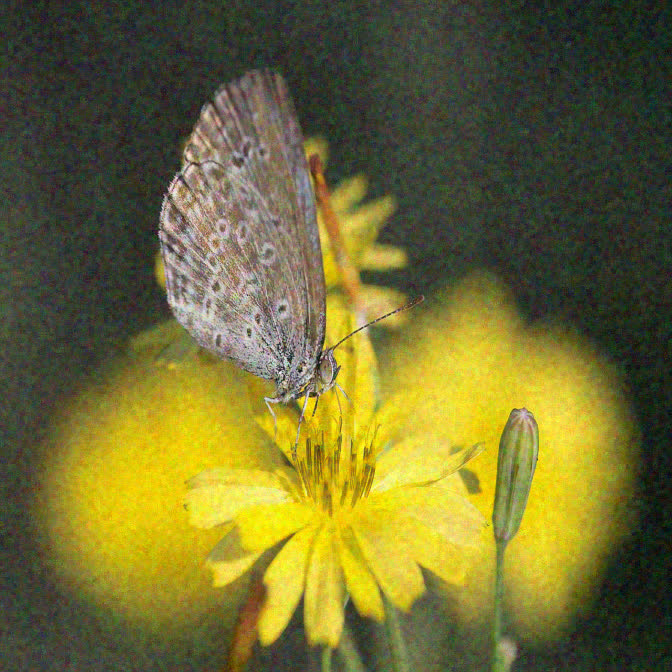}
    \caption{$\ell_2 = \mathbf{17.77}$}
\end{subfigure}
\hspace{0.1em}
\begin{subfigure}[t]{0.12\linewidth}
    \includegraphics[width=\linewidth]{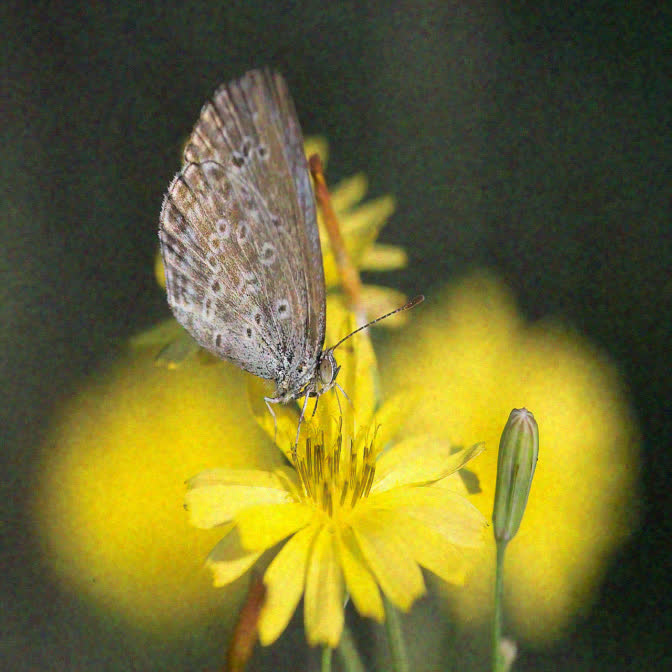}
    \caption{$\ell_2 = \mathbf{10.06}$}
\end{subfigure}
\hspace{0.1em}
\begin{subfigure}[t]{0.12\linewidth}
    \includegraphics[width=\linewidth]{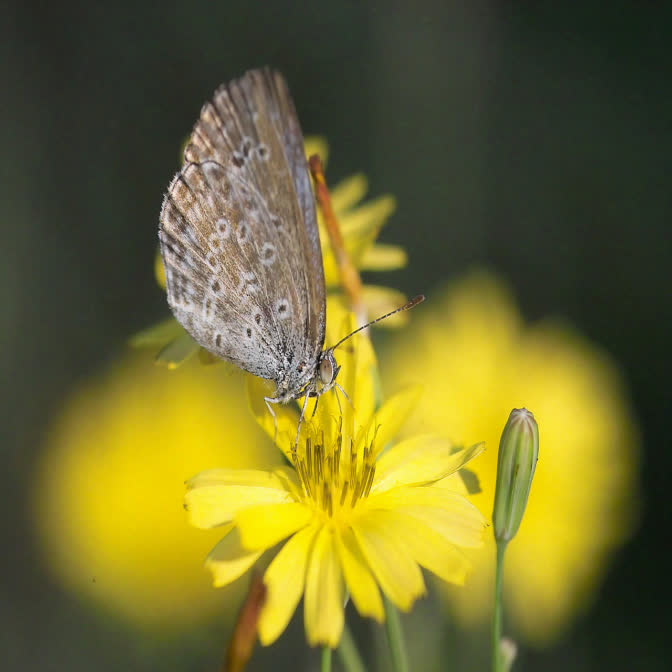}
    \caption{$\ell_2 = \mathbf{2.37}$}
\end{subfigure}
\hspace{0.1em}
\begin{subfigure}[t]{0.12\linewidth}
    \includegraphics[width=\linewidth]{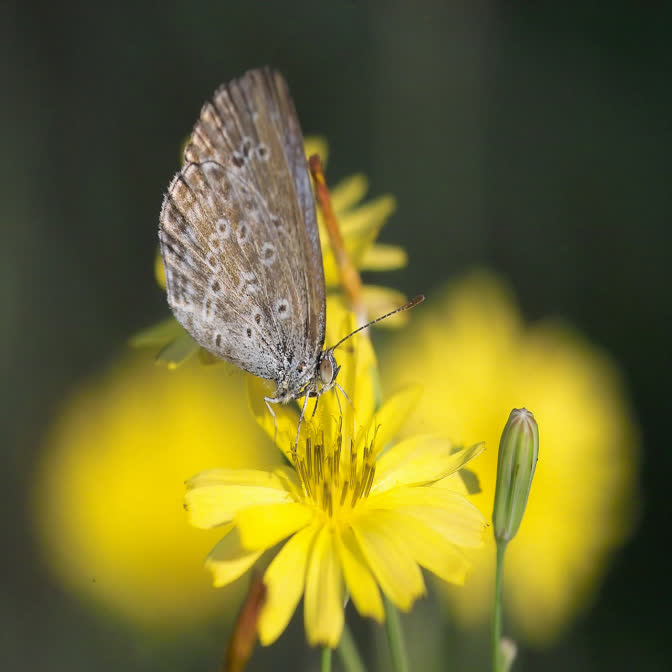}
    \caption{$\ell_2 = \mathbf{1.41}$}
\end{subfigure}

\caption{Adversarial examples in the black-box setting. 1st–3rd (4th–6th) rows: examples generated by HSJ, HR HSJ, and HR HSJ under the median filtering defense. 1st–6th columns: examples at 100, 200, 500, 1K, 5K, 10K model queries. Perturbations: the scaled \LL{2}-norm distance to the original image, numbers in bold font denote obtaining less perturbation than the LR HSJ attack. The shape of above images from LR and HR HSJ attacks is $224\times224$ and $672\times672$, respectively.}
\label{fig:attack:bb-more}
\end{figure*}

\Cref{fig:attack:bb-more} shows more black-box adversarial examples from our HR HSJ attack. HR HSJ attack is able to produce less perturbation than the LR HSJ attack for a given query budget. The same observation holds for median-defended scaling after 200 model queries.

\end{document}